\documentclass[12pt]{article}

\usepackage{graphicx}
\usepackage{caption}
\usepackage{subcaption}

\usepackage[usenames,dvipsnames]{color}
\usepackage{amssymb}
\usepackage[normalem]{ulem}
\usepackage{amsmath}
\usepackage{cite}
\usepackage{relsize}

\usepackage{cite}

\newcommand{\be}{\begin{equation}}
\newcommand{\ee}{\end{equation}}
\newcommand{\bea}{\begin{eqnarray}}
\newcommand{\eea}{\end{eqnarray}}

\def\Vector#1{\mbox{\boldmath $#1$}}

\def\vk{{\Vector k}}

\def\1{{\rm 1}}

\def\s1{^{\rm (1)}}

\DeclareMathAlphabet\mathbfcal{OMS}{cmsy}{b}{n}

\begin{document}

\baselineskip=17pt

\renewcommand{\thefootnote}{\fnsymbol{footnote}}

\begin{center}
\begin{Large}
{\bf Iterative Construction of Gaussian Process Surrogate}\\ 
{\bf Models for Bayesian Inference}\\
\end{Large}
\bigskip
\bigskip
Leen Alawieh$^1$, Jonathan Goodman$^2$, John B.~Bell$^3$\\

\medskip

$^1$Mechanical Engineering Department\\
American University of Beirut\\
Beirut 1107 2020, Lebanon
\\

\bigskip

$^2$Courant Institute of Mathematical Sciences \\
New York University\\
New York, NY 10012, USA\\

\bigskip

$^3$Center of Computational Science and Engineering \\
Lawrence Berkeley National Laboratory\\
Berkeley, CA 94720, USA\\

\end{center}

\vspace{3cm}

\begin{tabbing}
Corresponding Author: \hspace{5mm} \= Leen Alawieh \\
       \> Mechanical Engineering Department \\
       \> American University of Beirut\\
       \> Beirut 1107 2020, Lebanon \\
       \> \\
Phone: \> (+961)-1-350000 ext. 3457 \\
Email: \> la119@aub.edu.lb   \\
\\
To appear in: \> {\it Journal of Statistical Planning and Inference} \\
DOI: \>https://doi.org/10.1016/j.jspi.2019.11.002 \\
%\> \\
\\
Submitted: February 2019; \, Revised: October 2019; \, Accepted: November 2019\\

\end{tabbing}

\clearpage

\baselineskip=22pt

\section*{Abstract} 

A new algorithm is developed to tackle the issue of sampling non-Gaussian model parameter posterior probability distributions that arise from solutions to Bayesian inverse problems. The algorithm aims to mitigate some of the hurdles faced by traditional Markov Chain Monte Carlo (MCMC) samplers, through constructing proposal probability densities that are both, easy to sample and that provide a better approximation to the target density than a simple Gaussian proposal distribution would. To achieve that, a Gaussian proposal distribution is augmented with a Gaussian Process (GP) surface that helps capture non-linearities in the log-likelihood function. In order to train the GP surface, an iterative approach is adopted for the optimal selection of points in parameter space. Optimality is sought by maximizing the information gain of the GP surface using a minimum number of forward model simulation runs. The accuracy of the GP-augmented surface approximation is assessed in two ways. The first consists of comparing predictions obtained from the approximate surface with those obtained through running the actual simulation model at hold-out points in parameter space. The second consists of a measure based on the relative variance of sample weights obtained from sampling the approximate posterior probability distribution of the model parameters. The efficacy of this new algorithm is tested on inferring reaction rate parameters in a 3-node and 6-node network toy problems, which imitate idealized reaction networks in combustion applications. 
\\
\\
{\bf{Keywords}}: Gaussian process regression; active learning; surrogate models; Bayesian inference; MCMC.

\clearpage

\section{Introduction}
\label{sec:intro}

Mathematical models are constructed to approximate physical systems, which are then used to make predictions about their behavior at a given set of inputs. This constitutes solving the forward problem. On the other hand, inverse problems involve using observations in order to make inferences about the model inputs, or even inferences about the form of the models themselves. Posing the inverse problem in a Bayesian setting allows one to regularize any ill-posedness present, to account for any source of noise in the observations and prior uncertainty in the forward models, and to subsequently infer a posterior probability distribution for the inputs or models, as opposed to inferring a single best set of input values or a single best model~\cite{Iglesias2014, Tarantola2006, Tarantola2005}. The inferred posterior probability distribution summarizes all available information about the inputs or models, including a quantifiable measure of uncertainty that could, in turn, be propagated forward to provide a measure of uncertainty in the resulting predictions~\cite{Marzouk2007, Martin2012}. 

For most inverse problems of interest, there is no analytical representation of the posterior probability distribution, so statistical information about the distribution is typically extracted using Markov chain Monte Carlo (MCMC) sampling techniques, which entails solving the forward problem several times. As a result, standard MCMC methods are known to struggle when the underlying posterior distribution is complex, and when the forward model is computationally expensive~\cite{Marzouk2007, Martin2012}. There exist several approaches for tackling such challenges, some of which focus either on devising better sampling strategies~\cite{Martin2012, Morzfeld2016, Haario2006, Christen2005, Christen2010, Apte2007, Roberts1996, Dostert2006, Efendiev2006, Berger2003, Neal2011, Geweke1999, Girolami2011, Cui2016}, or on alleviating the cost of the forward model computations through developing reduced order models~\cite{Bui-Thanh2008a, Bui-Thanh2008, Arridge2006, Wang2005, Rozza2007, Ghanem1991, LeMaitre2010}, or cheaper surrogate representations that can (locally or globally) approximate the forward model~\cite{Habib2007, Heitmann2010, Narayanan2004, Higdon2008, Kennedy2001, Kennedy2006, Bilionis2013, Blight1975, Koehler1996, Morris1993, Oakley2002, OHagan1978, Conrad2016}. In this work, we focus on the surrogate representation approach.

The approximation to the forward model could be either deterministic or statistical in nature. 
For instance,~\cite{Gilks1995, Martino2015, Martino2018, Meyer2008, Cai2008, Shao2013} aimed at constructing adaptive polynomial or spline approximations of the target density using deterministic regression techniques, and utilized these approximations as proposal densities for MCMC sampling. On the other hand, studies such as~\cite{Busby2009, Martino2017, OÕHagan2006, Wang2011} aimed at constructing a probabilistic approximation of the forward model (known as an emulator), which they then used either for Bayesian optimization purposes or for conducting uncertainty and sensitivity analyses. While our underlying goal here is analogous to the former set of studies, the approach that we take is similar to the latter set of studies in that we implement a probabilistic, kernel-based regression method in order to approximate the target density and then seek to sequentially improve the approximation as further explained below in more details.

In this paper, we adopt the surrogate model approach, where we aim at constructing a faithful approximation of the posterior probability density using Gaussian process regression (GPR). Given that the forward model is an expensive black box computer simulation, we address the question of how to efficiently select data points for training the Gaussian Process (GP), such that we obtain a relatively accurate surrogate model using the minimum number of training points. To this end, we take an active learning approach, and iteratively build our surrogate surface. We develop a data selection criterion to decide, at each iteration stage, where in input space we should run the computer simulation next. Several previous studies have tackled a similar problem to the one presented here~\cite{Cohn1996, Cohn1996a, Cohn1997, Seo2000, MacKay1992, Gramacy2004, Gramacy2008, Gramacy2009, Gramacy2015, Seeger2003, Paass1995, Rasmussen2003, Kandasamy2017, Sacks1989, Currin1988, Currin1991, Preuss2018, Jones1998, Martino2017, Busby2009, Christen2011}, however in most of these studies, the data selection criterion implemented is either not appropriate for our current purposes due to its associated computational expense, or is coupled with an unnecessarily complicated GP model. For example, in~\cite{Cohn1996a, Cohn1997, Seo2000, MacKay1992, Gramacy2004}, the selection criteria aim to minimize the predictive error (or simply the predictive variance) of the surrogate model, but they require estimating integrals over the entire input space or computing local sensitivity derivatives, which could be computationally expensive or not even readily feasible. 
Rasmussen~\cite{Rasmussen2003} takes a similar approach to ours, in that he couples an MCMC sampler (specifically, Hamiltonian Monte Carlo sampler) with a GP model, in order to obtain potential data point candidates for training the GP. However, he offers no mathematical justification to show how he arrived to the specific selection heuristic used for choosing the data point candidates. Moreover, Rasmussen's algorithm seems to require computing first and second order covariance derivatives for optimizing the GP model, as well as computing the forward model and its partial derivatives several times at each iteration stage. On the contrary, beyond solving the forward model at the input point selected to add to the training set, the algorithm we develop in this paper does not require solving the forward problem or any of its derivatives while seeking potential data point candidates. In addition to that, no optimization of the GP model is required at any stage of the algorithm. In fact, we will demonstrate that, despite choosing a naive unoptimized GP model, our algorithm can still construct a faithful surrogate model. Kandasamy et al.~\cite{Kandasamy2017} suggest the same data selection criterion as the one we derive here, however in order to locate the optimal data point that maximizes the criterion, they evaluate the criterion on a coarse grid in input space, which becomes impractical in high dimensions. We circumvent this hurdle by coupling our GP model with an MCMC sampler, which allows us to cheaply locate the optimal data point.

This paper is organized as follows. In Section~\ref{sec:formu} we start by briefly introducing Bayesian inference and Gaussian process regression, and then move on to develop the algorithm for the optimal selection of GP training data. Implementation of the developed algorithm on a number of network toy problems of increasing dimensionality is presented in Section~\ref{sec:res}. Major conclusions are finally summarized in Section~\ref{sec:conc}.

\bigskip\bigskip
\section{Problem Formulation} 
\label{sec:formu} 

\subsection{Bayesian parameter inference}

We consider a forward problem defined by:
\be 
\Vector y = \mathbfcal M  ( \Vector \theta ) + \Vector \eta  
\label{eq:forward model} 
\ee 
where $\Vector y$ is a real-valued $d$-dimensional vector of observed data, $\mathcal M$ is a forward model that is a function of a real-valued $n$-dimensional vector of parameters $\Vector \theta$, such that $\mathcal M: \mathbb R^n \rightarrow \mathbb R$. It is assumed that $\Vector \theta$ is unknown or uncertain, and is thus treated as a vector of real-valued random variables. $\Vector \eta$ is a $d$-dimensional vector of i.i.d.\ random variables with a probability density p$_\eta$, which accounts for the discrepancy between the model predictions and the observed data due to measurement errors only. Note that, as expressed above, it is assumed that the noise is additive. 

Given the observed data, our goal is to statistically infer the model parameters, $\Vector \theta$, using the Bayesian approach. According to Bayes' theorem, the solution to this inverse problem is given by: 
\be  
p(\Vector \theta \vert \Vector y) \propto p(\Vector y \vert \Vector \theta) \, p(\Vector \theta) 
\label{eq:Bayes rule} 
\ee    
\\ 
where $p(\Vector \theta \vert \Vector y)$ represents the posterior probability distribution of $\Vector \theta$ after knowledge from the observed data has been incorporated. $p(\Vector \theta)$ represents the prior probability distribution of the model parameters, which encodes knowledge about the parameters before the data has been observed. $p(\Vector y \vert \Vector \theta)$ represents the likelihood function, which describes the probability that a given set of model parameters, $\Vector \theta$, gives rise to the observed data, $\Vector y$. Using our assumptions about the distribution of $\Vector \eta$, the likelihood function can be written as: 
\be 
L (\Vector \theta) \equiv p(\Vector y \vert \Vector \theta) = p_\eta \left ( \Vector y - \mathbfcal M  ( \Vector \theta ) \right )  
= \prod_i p_\eta \left ( y_i - \mathcal M ( \Vector \theta) \right )  
\label{eq:likelihood function}
\ee 
\\ 
If we further assume that the noise, $\Vector \eta$, is zero mean Gaussian with a scalar variance, $\beta$, and that the observed data are independent and identically distributed, then the likelihood function becomes of the form:
\be
L (\Vector \theta) =   \prod_i \mathcal N \left (\mathcal M ( \Vector \theta), \beta \right )
\label{eq:likelihood Gaussian}
\ee

When the forward model $\mathcal M$ is nonlinear, as is typical in most applications, an analytical solution of Eq.~(\ref{eq:Bayes rule}) becomes intractable. Moreover, if the dimensionality, $n$, of the vector of model parameters, $\Vector \theta$, is high ( $> 2-3$), and the forward model is expensive to solve, then a simple grid-based numerical solution of the Bayesian inverse problem becomes prohibitively expensive. In these cases, one usually resorts to a statistical characterization of the posterior probability distribution of $\Vector \theta$ through random sampling techniques, such as Markov Chain Monte Carlo (MCMC)~\cite{Tarantola2005}. However, for the statistical quantities obtained through random sampling to converge to the true ones, the MCMC sampler will need be run for a long time, which consequently requires multiple evaluations of the likelihood function. This is especially true for complicated, highly non-Gaussian posterior distributions~\cite{Martin2012, ElMoselhy2012, Parno2018}. Since each evaluation of the likelihood function involves a forward solve of the physical model, direct random sampling could become computationally impractical (or even infeasible), if the physical model is computationally expensive. One way to circumvent this hurdle is to replace the likelihood function with a cheaper surrogate model which faithfully reflects the likelihood response surface in regions of parameter space that are associated with high probabilities. We choose to employ a kernel method, specifically the Gaussian process regression (GPR) technique, for constructing such a surrogate model, mainly due to its generality and flexibility. 

\subsection{Gaussian process regression}
\label{sec:GPR}

Gaussian process regression is a non-parametric method that stems from Bayesian linear regression. The latter tends to define a probability distribution over the weighting parameters of a given set of nonlinear basis functions, whereas the former defines a probability distribution over the regression functions directly. Specifically, given a set of $N$ observations $\Vector L = \{ L_1, \ldots, L_N \}$ at a set of input points $\Vector \Theta = \{ \Vector \theta_1, \ldots, \Vector \theta_N \}$, a Gaussian process defines a probability distribution over functions, $L(\Vector \theta)$, such that the joint probability distribution of the $N$ observations, $\Vector L$, is Gaussian with a mean $\Vector \mu(\Vector \theta) = \mathbb E[\Vector L(\Vector \theta)]$ and a covariance $\Vector K_N = \mathbb E[(\Vector L(\Vector \theta) -  \Vector \mu(\Vector \theta) )(\Vector L(\Vector \theta') -  \Vector \mu(\Vector \theta'))^T ]$. $\Vector K_N$ is an $N \times N$ matrix whose elements are composed of the kernel function $k(\Vector \theta, \Vector \theta')$. 

Assuming no prior knowledge about the mean, we will take $\mu (\Vector \theta) = 0$. There are several options that one could choose for the covariance kernel function~\cite{Rasmussen2006}, but in what follows, we will focus on the stationary, isotropic squared-exponential kernel function given by:
\be
k(\Vector \theta, \Vector \theta') = s^2 \text{exp} \left ( - \, \frac{1}{2} \, \mathlarger {\sum}\limits_{i = 1}^{n} \frac{(\theta_i-\theta_i')^2}{ \ell^{2}} \, \right )
\label{eq:sqexp_kernel}
\ee
\\
where $s^2$ is the variance (diagonal component of $\Vector K$), and $\ell$ is a correlation length-scale parameter. Typically, optimal values for the kernel parameters are determined by optimizing the log-marginal likelihood of the Gaussian process using standard techniques such as the conjugate gradient method~\cite{Rasmussen2006, MacKay1998}. 

Given that the joint distribution of the outputs at a given set of inputs is normal, then the conditional predictive distribution at a new input point, $p(L_{N+1} | \Vector L)$, is accordingly also Gaussian with a mean and variance given by:
\bea
\mu(\Vector \theta_{N+1}) &=& \vk^T \Vector K^{-1}_N \Vector L
\label{eq:mean_pred}
\\
\sigma^2(\Vector \theta_{N+1}) &=& k(\Vector \theta_{N+1}, \Vector \theta_{N+1}) - \vk^T \Vector K^{-1}_N \vk
\label{eq:var_pred} 
\eea  
where $\vk$ is an $N \times 1$ kernel vector composed of the kernel function evaluated at $\Vector \theta_{N+1}$ and each of the  $N$ training input points in $\Vector \Theta$. 
This gives us the predictive mean and the associated uncertainty at a new test input point, given observations at previous training input points. 

\subsection{Optimal selection of training data}

Our aim is to build a surrogate model for the likelihood function in Eq.~(\ref{eq:likelihood function}), given that the forward model, $\mathcal M(\Vector \theta)$, is computationally expensive. Thus, it is important to judiciously choose the data for training the GP surface, such that we get a faithful surrogate representation of the likelihood function with the fewest number of training data. 
It has been shown in numerous studies~\cite{Kandasamy2017, Fedorov1972, Seo2000, Cohn1996, Cohn1996a, Cohn1997, Paass1995} that an ``active learning" or ``sequential design" approach to the problem of optimal data selection is more efficient than ``passive learning" strategies, and this is the approach that we opt to adopt in this paper. 

Assuming that our choice of the GP covariance kernel function is suitable, and that we have initialized our GP surface with $N$ initial training input-output data pairs $\{ (\Vector \theta_1, L_1), \ldots, (\Vector \theta_N, L_N) \}$, our task is to determine which data point we should select to add to our training data set such that it maximizes the amount of information gained by our GP surface. In other words, we wish to determine the optimal $\Vector \theta_{N+1}$ at which to evaluate our likelihood function next. Subsequently, the new data point $(\Vector \theta_{N+1},L_{N+1})$ is added to our training data set, the GP surface is updated, and the process is repeated until the accuracy of the resulting GP surface is deemed satisfactory. Note that we will not be concerned in this study with the issue of selecting multiple new observation points at once --- this will be the subject of future studies. 

\subsubsection{Utility measure}
\label{sec:utility measure}

In order to determine the potential amount of information that a given data point carries, we need to define a utility measure that could serve as an information gauge. For a GP surface, each point in input space is associated with a mean and a variance (uncertainty) around that mean. Thus, a data point that would maximize the amount of information gain, would be the one that results in a maximum reduction of uncertainty at a given point in input space. In this case, a possible information gauge would be the point-wise predictive variance given by Eq.~(\ref{eq:var_pred}). However, recall that our ultimate goal is to create an approximation of the likelihood surface which we could then use for MCMC sampling. As a result, we are only interested in an accurate approximation of the regions in $\Vector \theta$ space that are associated with a high posterior probability. On the other hand, regions that have a low posterior probability are not important for statistically characterizing the posterior probability distribution of $\Vector \theta$, and thus do not require an accurate representation by our GP surrogate surface. Consequently, our criterion should be able to weight the point-wise variance with the corresponding posterior probability value at a given input point.  

In order to derive a probability-weighted point-wise variance, we start out by writing the point-wise variance according to our GP-approximated surface. Let $\pi(\Vector \theta)$ be the true posterior probability at a given input point, and $\widetilde{\pi}(\Vector \theta)$ its corresponding GP approximation (Note that, as will be noticed in the equations that follow, the GP model by itself approximates the log-likelihood). Accordingly, 
\be
\widetilde\pi(\Vector \theta) = \frac{1}{z} \, e^{\overline{\phi}\left(\Vector \theta\right)} \, e^{-w\left(\Vector \theta\right)}       
\label{eq:GP_post}
\ee
\\
where $z$ is a normalization factor, $\overline \phi = \mu_{\text{prior}} + \mu_{\text{GP}}$, and $w \sim \mathcal N(0,\sigma^2)$. $\mu_{\text{GP}}$ is the GP predictive mean, and $w$ is the associated GP uncertainty. $\mu_{\text{prior}}$ accounts for any prior information or assumptions about the distribution of $\Vector \theta$. For example, if our prior assumption is that $\Vector \theta \sim \mathcal N(\Vector m_\theta, \Vector \Sigma_\theta)$, then 
$$
\pi_\text{prior}(\Vector \theta) = \text{det}(2 \pi \Vector \Sigma_\theta)^{-1/2} \, \text{exp} \left( -\frac{1}{2}(\Vector \theta - \Vector m_\theta)^T \, \Vector \Sigma_{\theta}^{-1} \, (\Vector \theta - \Vector m_\theta) \right)
$$
and $\mu_{\text{prior}} = \text{log} \left( \pi_\text{prior}(\Vector \theta) \right)$.
\\  
The variance of $\widetilde{\pi}(\Vector \theta)$ is given by:
\bea
var (\widetilde\pi(\Vector \theta)) &=& \frac{1}{z^2} \, var \left( e^{\overline{\phi}\left(\Vector \theta\right)} \, e^{-w\left(\Vector \theta\right)} \right) \nonumber 
\\ [10pt]
&=& \frac{1}{z^2} \, e^{2\overline{\phi}\left(\Vector \theta\right)} \, var \left( e^{-w\left(\Vector \theta\right)} \right) 
\\ [10pt]
&=& \frac{1}{z^2} \, e^{2\overline{\phi}} \, \left( \mathbb E \left[ e^{-2w} \right] - \mathbb E \left[ e^{-w} \right]^2 \right) \nonumber
\eea
where we omitted the dependence on $\Vector \theta$ in the last expression to simplify the notation. Using simple calculus, it can be shown that $\mathbb E \left[ e^{-w} \right] = e^{\sigma^2 / 2}$. This leads to:
\be
var (\widetilde\pi(\Vector \theta)) \propto e^{2\overline{\phi}} \, \left( e^{2 \sigma^2} - e^{\sigma^2} \right)
\label{eq:var_approx}
\ee
The above equation gives us our sought after point-wise probability-weighted predictive variance, which we will employ as our criterion for determining the next best data point to add to our training data set. In other words, we will choose to evaluate $L_{N+1}$ at the $\Vector \theta_{N+1}$ that maximizes Eq.~(\ref{eq:var_approx}), and add this new data point, $(\Vector \theta_{N+1},L_{N+1})$, to our GP training data set. If Eq.~(\ref{eq:var_approx}) happens to admit multiple maxima, then we will randomly pick a $\Vector \theta_{N+1}$ out of the set of possible maxima.

\subsubsection{Searching for the optimal point} 
\label{sec:optimal point}

Having derived our measure of optimality using the variance of the approximated posterior probability density, we now need a way to locate the point that maximizes Eq.~(\ref{eq:var_approx}), especially when our input space is high-dimensional. Since the surface that we are seeking to optimize could potentially be multimodal and/or non-smooth, common optimization techniques would not be the best resort as they are known to struggle in such cases. Instead, we will seek to locate our optimal point via MCMC sampling. Specifically, we will sample the surface given by Eq.~(\ref{eq:var_approx}) using the emcee implementation of the affine invariant ensemble MCMC sampler~\cite{Goodman2010, Foreman-Mackey2013}, which we will hereafter refer to as the Hammer sampler. 
(We should note that the use of Hammer is not necessary. Any other MCMC sampler could be employed, however, we chose to use Hammer because of its efficiency in sampling highly non-Gaussian surfaces.)
From the set of samples obtained from Hammer, we pick the sample point with the highest probability-weighted variance as given by Eq.~(\ref{eq:var_approx}). Note that, in order to prevent the covariance kernel matrix $\Vector K$ from becoming ill-conditioned, we also require that our potential new data input point, $\Vector \theta_{N+1}$, be a certain distance from the other input points that are already in the training set $\Vector \Theta$. We found that a constraint on the Eucledian norm given by: $$|| \Vector \theta_{N+1} - \Vector \theta_i ||_2 > 0.2\, \ell \quad \forall \, \Vector \theta_i \in \Vector \Theta$$ where $\ell$ is the correlation length-scale defined in Eq.~(\ref{eq:sqexp_kernel}), suffices for this purpose. If our potential $\Vector \theta_{N+1}$ sample point violates this condition, then we pick instead the next best sample point that satisfies this condition. 
Note that the above distance criterion is needed due to the potential incompatibility between the smoothness imposed by the underlying covariance kernel function and that of the data. Had we opted for optimizing the GP model, then this condition would no longer be necessary. Moreover, a factor different than 0.2 could have in principle been chosen, but we picked it as a reasonable value that would not sacrifice flexibility for smoothness.

\subsubsection{Accuracy measure}
\label{sec:accuracy measure}

As mentioned before, we are aiming for constructing a faithful GP-surrogate of the likelihood function. Thus, we need a measure by which we can judge the accuracy of our approximation at each iteration stage. For this purpose, we will rely on two different quality measures. The first quality measure is based on the empirical average absolute error between predictions obtained from the approximate surface and predictions obtained from the true surface at a given set of hold-out points in parameter space. The hold-out points are obtained by running the Hammer sampler on both, the approximate and the true surfaces, and using the sample points after burn-in as our hold-out points. We resort to two sets of samples so that we do not bias our quality measure towards sample points whose true likelihood is either high or small (in other words, we would like to check the quality of our approximate surface in areas where the true likelihood is large, and also in areas where the true likelihood is small but is incorrectly classified as large by our approximation).
Accordingly, this gives us two error measures; one weighted by the true probability distribution, $\pi (\Vector \theta)$, and the other weighted by the approximate probability distribution $\widetilde{\pi} (\Vector \theta)$. We designate these as $\mathcal E_\text{true}$ and $\mathcal E_\text{approx}$, respectively, and they are of the form:
\be
\mathcal E = \frac{1}{N} \sum \limits_{i = 1}^{N} \, \bigl\lvert \text{log}\left( \widetilde{\pi}(\Vector \theta_i) \right) - \text{log}\left( \pi(\Vector \theta_i) \right) \bigl\rvert
\label{eq:abs_error}
\ee
\\
where $N$ is the number of samples generated by the Hammer sampler. 

The second quality measure, $R$, is based on the relative variance of sample weights, $\rho$, which is computed as follows~\cite{Morzfeld2016, Vanden-Eijnden2013}:
\bea
w &=& \frac{\pi (\Vector \theta)}{\widetilde{\pi} (\Vector \theta)} \quad ; \quad \rho = \frac{var\left[ w \right]}{\mathbb{E}\left[ w \right]^2}  \nonumber \\
\nonumber \\
R  &=& \rho + 1 = \frac{\mathbb{E}\left[ w^2 \right]}{\mathbb{E}\left[ w \right]^2}
\label{eq:Rmeasure}
\eea
\\
where $w$ is the weight of a sample point generated from Hammer by sampling the surrogate $\widetilde{\pi} (\Vector \theta)$ surface. Note that $R \ge 1$, and that $R = 1$ when $\widetilde{\pi} (\Vector \theta) = \pi (\Vector \theta)$ (in which case, the surrogate surface is an exact representation of the true surface). One has to be careful, though, that the mean of the sample weights is close to 1. Otherwise, the variance of the weights could still be zero, even though the surrogate surface is too off from the true surface.
This happens if the ratio of $\pi (\Vector \theta)$ to $\widetilde{\pi} (\Vector \theta)$ is some constant, in which case, $\widetilde{\pi} (\Vector \theta)$ would need to be re-scaled by normalization with this constant factor.

\bigskip\bigskip
\section{Implementation}
\label{sec:res}

In what follows, we will test the efficacy of our training algorithm on $n$-node network toy problems of increasing dimensionality. The main motivation behind this study is the inference of reaction rate parameters in combustion problems, so the toy problems are meant to imitate an idealized reaction network model for the reaction kinetics in combustion applications.      
For each problem, we start first by running the Hammer sampler on the true posterior distribution  $\pi_{\text{post}}(\Vector \theta)$, in order to obtain information about the mean and the covariance of the underlying posterior surface. The number of walkers used by the Hammer sampler is twice the dimensionality of the problem, and the walkers are initialized by drawing samples from a Gaussian distribution. 
We then use this information on the sample mean and covariance to construct a Gaussian prior distribution, $\pi_\text{prior}(\Vector \theta) = \mathcal N(\Vector m_\theta, \Vector \Sigma_\theta)$. Note that this initial Gaussian approximation does not give us complete information about the posterior surface, since all of our test models are non-linear, which makes the posterior distribution non-Gaussian by construction. Our aim is to test whether our GP-augmented surrogate surface can help resolve the non-linearities that a Gaussian proposal fails to capture. Moreover, it is not necessary to run the sampler on the true surface for a long time. It is sufficient to obtain only preliminary information about the mean and covariance of the true distribution, as long as the support of our initial Gaussian approximation is large enough so as not to preclude high probability regions that the sampler did not manage to visit during the preliminary sampling stage. This could be achieved, for example, by inflating the sample variance used in constructing the Gaussian prior by some factor. 

We will assume that the true likelihood function has a Gaussian form, which leads to the GP likelihood surrogate to be given by: 
\be
\widetilde{\pi}_\text{lik} \left( \Vector \theta \right) = e^{\, \mu_\text{GP} \left( \Vector \theta \right)}
\label{eq:lik_GP}
\ee 
\\
Consequently, our GP-augmented surrogate for the posterior distribution becomes:
\bea
\widetilde{\pi}_\text{post} \left( \Vector \theta \right) &\propto& \widetilde{\pi}_\text{lik}\left( \Vector \theta \right) \, \pi_\text{prior}\left( \Vector \theta \right) \nonumber \\
&=& e^{\, \mu_\text{GP} \left( \Vector \theta \right)} e^{\, \mu_\text{prior} \left( \Vector \theta \right)}
\label{eq:post_GP}
\eea
where, as mentioned in Section~\ref{sec:utility measure}, $\mu_\text{prior} \left( \Vector \theta \right) = \text{log} \left( \pi_\text{prior}(\Vector \theta) \right)$. $\mu_\text{GP} (\Vector \theta)$ is given by Eq.~(\ref{eq:mean_pred}) after initializing the GP surface with a certain number of training points, while $\mu_\text{GP} (\Vector \theta) = 0$ before adding any training points, and $L(\Vector \theta)$ represents an observation of the true log-likelihood function, $\text{log}(\pi_\text{lik}(\Vector \theta))$. 
Note that comparing Eq.~(\ref{eq:post_GP}) above with the previous Eq.~(\ref{eq:GP_post}), only the GP predictive mean is included above without the GP variance. The GP variance was used to derive the sought after utility measure, but for prediction purposes (in this case, to predict the value of the posterior surrogate at a given input $\Vector \theta$), only the mean is needed. The GP variance is needed only as a way to assess the uncertainty or confidence in our posterior approximation.

For all of the test problems presented below, we will adopt the isotropic squared-exponential kernel given by Eq.~(\ref{eq:sqexp_kernel}) as our covariance kernel function. We will initialize the covariance kernel parameters to be $s^2 = max\{ | \text{log}(\pi_\text{lik}(\Vector \theta)) | \}$ and $\ell = 0.5$, and keep them fixed thereafter. The maximum absolute value of the true log-likelihood function is based on observations of log-likelihood values for samples proposed by Hammer (for both, the samples that were eventually accepted and those that were rejected) when sampling the true posterior in order to construct the Gaussian prior distribution, as described above. The value of $\ell$ chosen is based on the fact that it is usually a good initial guess when no prior knowledge of the correlation length-scales is available --- it is neither too small nor too large.  
We will not attempt to optimize or update the kernel parameters as we add GP training points, as one of the goals of this preliminary study is to demonstrate that it is not necessary to optimize the kernel parameters in order for our training algorithm to work. Of course, one expects the efficiency of the training algorithm to improve when optimal kernel parameters are implemented rather than non-optimal ones, but then the enhanced efficiency comes at the extra computational cost and added complexity of optimizing for the kernel parameters. A comparison between the computational cost and complexity of implementing an isotropic kernel with fixed parameters, an isotropic kernel with optimized parameters, and an anisotropic kernel with optimized parameters will be the subject of future investigations.

\subsection{3-node network model}
\label{sec:3-node network}

\begin{figure}[h]
\centering
\includegraphics[scale=0.7]{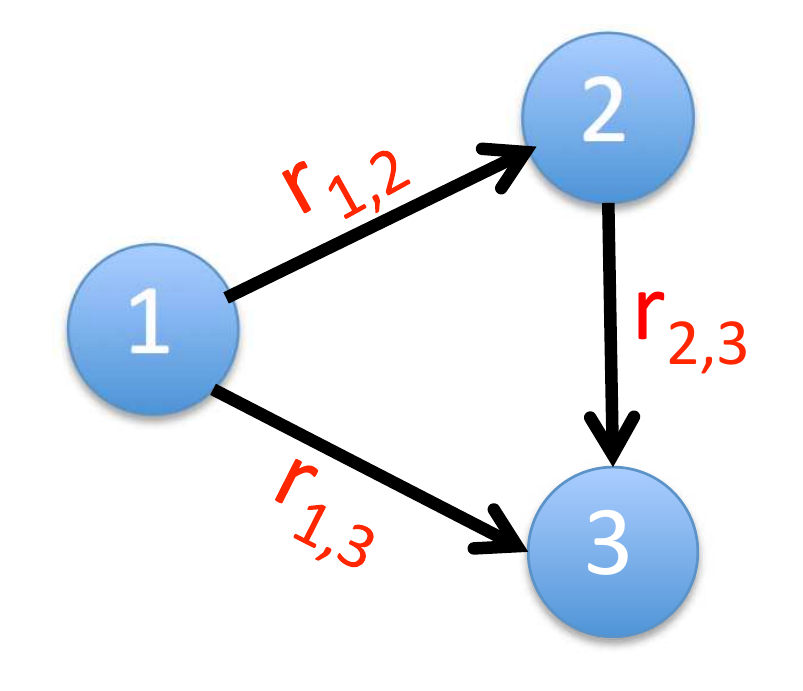}
\caption{Schematic of the 3-node network model.}
\label{fig:3-node_schematic}
\end{figure}

We start by considering a 3-node network model as illustrated by the schematic in Fig.~\ref{fig:3-node_schematic}. The reaction rate, $r$, across each of the nodes is given by:
$$ r_{i,j;k} = \text{C}_{i,j; k} \, A_{i,j} \, e^{-\beta_k E_{i,j}} $$
where $A$, $E$, $\beta$, and C are the pre-exponential factor, activation energy, thermodynamic inverse temperature, and concentration of the reaction species, respectively. The $i,j$ indices refer to the nodes across which a reaction is taking place, while the $k$ index corresponds to the specific experimental conditions under which the reactions occur. It is assumed that the reactions are irreversible, and that they proceed sequentially across each of the nodes with the first node always being 1 and the terminal node being the highest number node in the network (3 in this case). The eventual observed end-output of each experiment corresponds to the total time it takes the slowest reaction pathway to complete (similar to a rate limiting step in an actual reaction mechanism). The time for a given reaction pathway is given by the sum of the inverse of the reaction rate across each of the nodes involved in that pathway. For example, the total time, $t$, of the reaction pathway $\{ 1 \rightarrow 2 \rightarrow 3 \}$ is $t = 1/r_{1,2} + 1/r_{2,3}$.  

%%%%%%%%%%%%%%%%%%%%%%%%%%%%%%%% 

%%%%%%%%%%%%%%%%%%%%%%%%%%%%%%%% 
\subsubsection{2-D case}
\label{sec:2D model}

We start out with a 2-D scenario, in which we assume that only two of the activation energies, $ \Vector \theta = \{ E_{1 \rightarrow 2}, E_{2 \rightarrow 3} \}$, are uncertain with all other parameters being known. For inferring the uncertain parameters, we rely on observations from a set of 6-7 synthetic noisy experiments, not all of which are equally informative. The synthetic noisy observation data is obtained by numerically evaluating the model for each experiment using the true reaction parameters, and then adding a $\sigma$ factor of Gaussian noise, $\sim \sigma \, \mathcal N (0, 1)$, to the computed output from each experiment. For the 3-node network model, we used a noise level of $\sigma = 0.1$. The true network reaction rate parameters are shown in Table~\ref{tab:3node_params}, and Table~\ref{tab:3node_exps} lists the conditions for each experiment. 
%%%%%%%%%%%%%%%%%%%%%%%%%%%%%%%% 
\begin{table}[h]
\begin{center}
\begin{tabular} {| *{4}{c} |}
\hline
Nodes & (1,2) & (1,3) & (2,3)  \\ \hline
$E$ & 5 & 2 & 1  \\ \hline
$A$ & 1 & 3 & 2  \\
\hline
\end{tabular} 
\end{center}
\caption{True parameters of the 3-node network model. The rate of the reaction across each node is given by $r_{i,j;k} = \text{C}_{i,j; k} \, A_{i,j} \, e^{-\beta_k E_{i,j}}$.}
\label{tab:3node_params}
\end{table}

\begin{figure}[h!]
\centering
\includegraphics[scale=0.5]{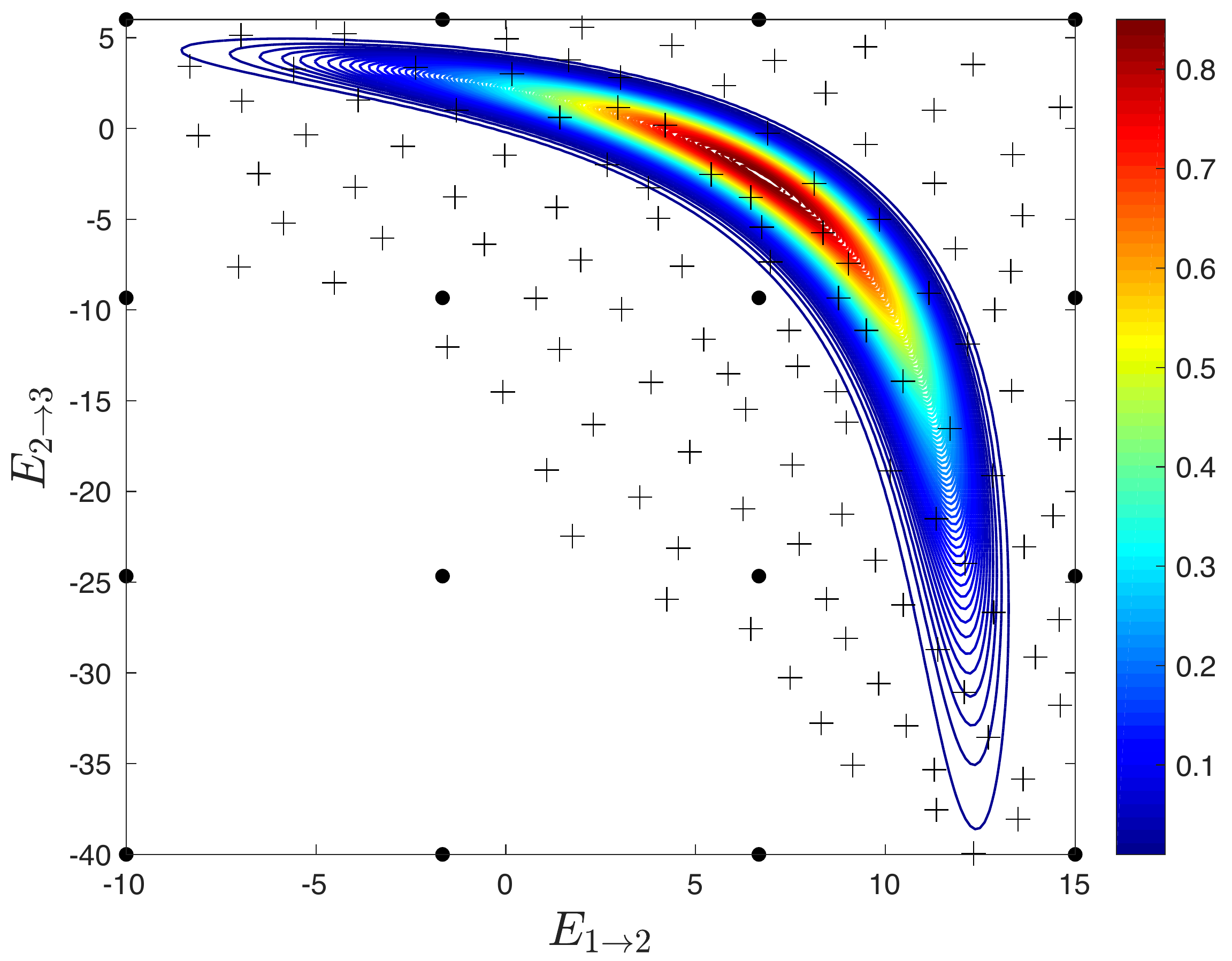}
\caption{Unnormalized true 2-D posterior distribution of $ \Vector \theta = \{ E_{1 \rightarrow 2}, E_{2 \rightarrow 3} \}$ for the 3-node network problem, given a Gaussian prior and observed data from 6 experiments. The black dots represent the initial training data points used for initializing the GP surrogate surface, and the black crosses represent the additional training points selected using the search algorithm.}
\label{fig:truedist-6exp}
\end{figure}

\begin{figure}[h!]
  \begin{subfigure}[t]{0.5\textwidth}
  \includegraphics[scale=0.35]{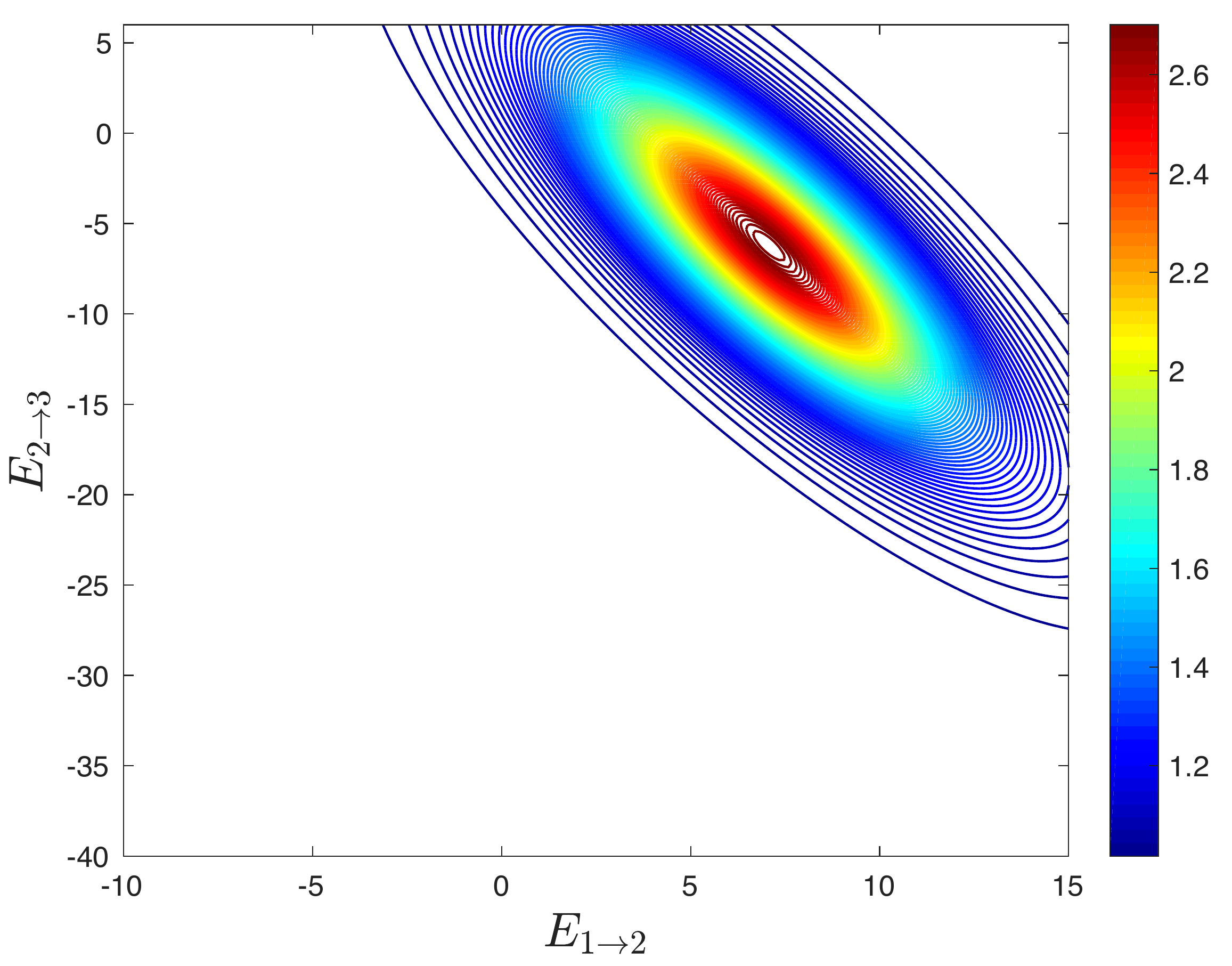}
  \caption{}
  \label{fig:Gauss-6exp}
  \end{subfigure}
  \begin{subfigure}[t]{0.5\textwidth}
  \includegraphics[scale=0.35]{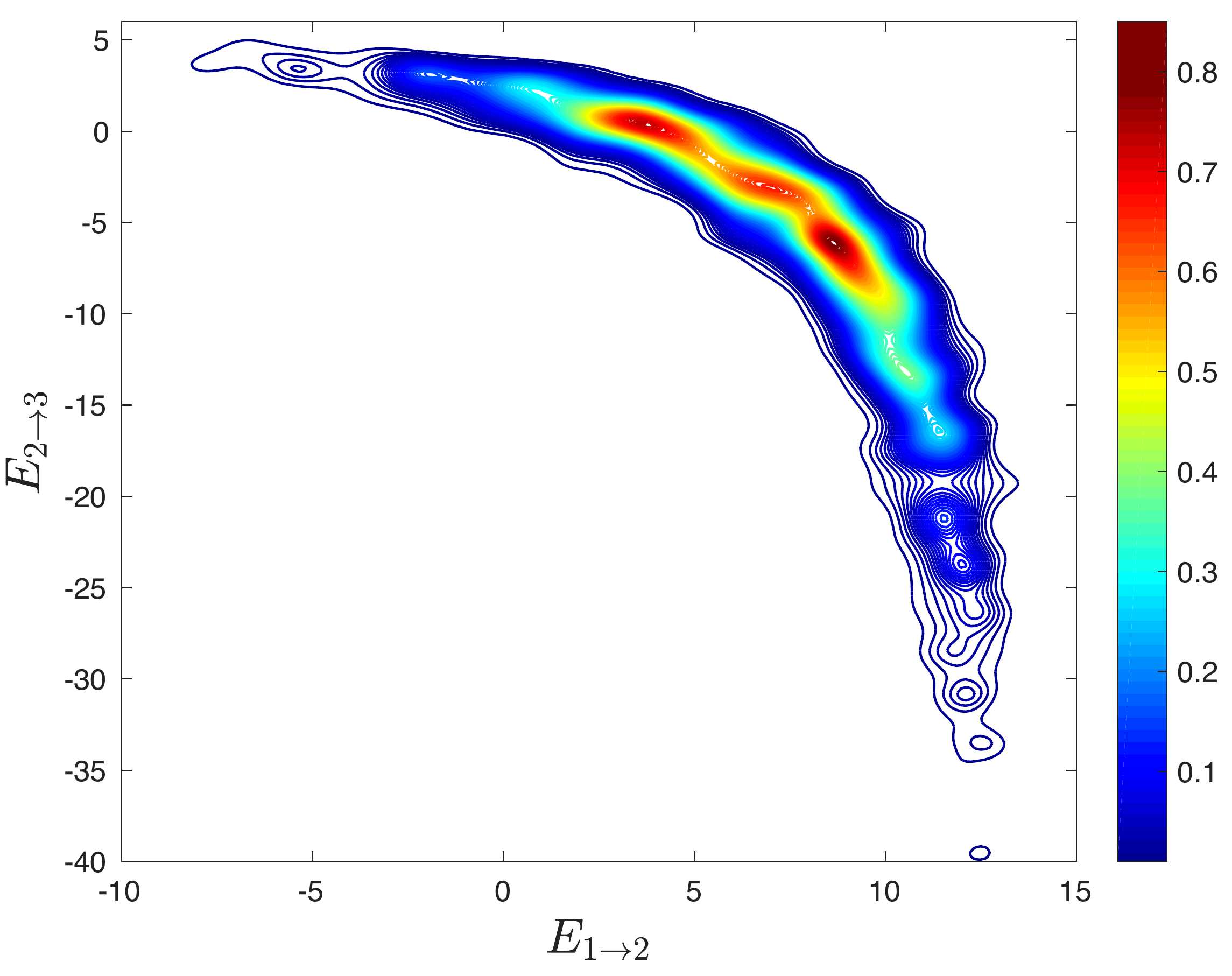}
  \caption{}
  \label{fig:GP200-6exp}
  \end{subfigure}
\caption{ (a) Unnormalized Gaussian prior distribution, $\pi_\text{prior}(\Vector \theta)$ (b) Surrogate posterior distribution, $\widetilde{\pi}_{post}(\Vector \theta)$, constructed using the 216 training points marked in Fig.~(\ref{fig:truedist-6exp}). }
\label{fig:surrogate-6exp}
\end{figure}

\begin{table}[h]
\begin{center}
\begin{tabular} {| c | *{4}{c} |}
\hline
Experiment $k$ & $\text{C}_{1 \rightarrow 2}$ & $\text{C}_{1 \rightarrow 3}$ & $\text{C}_{2 \rightarrow 3}$ & $\beta$ \\ \hline
1 & 10 & 0.5 & 10 & 0.01 \\ \hline
2 & 20 & 0.5 & 20 & 0.1  \\ \hline
3 & 0.5 & 20 & 2 & 0.01  \\ \hline
4 & 2 & 30 & 0.5 & 0.1  \\ \hline
5 & 2 & 20 & 0.5 & 0.01  \\ \hline
6 & 5 & 20 & 5 & 0.01  \\ \hline
7 & 0.5 & 30 & 0.5 & 0.1  \\ 
\hline
\end{tabular} 
\end{center}
\caption{Experiments used for inferring the uncertain reaction rate parameters in the 3-node network model.}
\label{tab:3node_exps}
\end{table}
%%%%%%%%%%%%%%%%%%%%%%%%%%%%%%%% 
Fig.~(\ref{fig:truedist-6exp}) shows the unnormalized true 2-D posterior distribution, $\pi_\text{post}(\Vector \theta)$, of the uncertain rate parameters given data observed using the first 6 experiments shown in Table~\ref{tab:3node_exps}. The distribution was obtained by numerically evaluating the true posterior on a $200 \times 200$ grid. Note that the Gaussian prior used in evaluating the true posterior distribution is not the same as the Gaussian prior, $\pi_\text{prior}(\Vector \theta)$, used to construct the surrogate posterior. To test whether we can recover this 2-D probability distribution using our sequentially trained GP-augmented surrogate surface, we started by constructing our Gaussian prior distribution and initializing the GP with 16 training points on a $4 \times 4$ grid. The initial training points used are shown as black dots in Fig.~(\ref{fig:truedist-6exp}). The initial Gaussian prior distribution is shown in Fig.~(\ref{fig:Gauss-6exp}), while Fig.~(\ref{fig:GP200-6exp}) shows the final $\widetilde{\pi}_\text{post}(\Vector \theta)$ surface, after sequentially training the GP surface with 200 additional observation points using the search algorithm described in Section~\ref{sec:optimal point}. The additional observation points selected are marked as black crosses in Fig.~(\ref{fig:truedist-6exp}). As can be seen from the figures, our training algorithm is able to select observation points in areas where the true posterior probability is high and also in areas that the original Gaussian prior incorrectly classified as important, resulting finally in a surrogate surface that correctly captures the nonlinear curvature of the true distribution. Notice also that despite using an isotropic covariance kernel whose parameters have not been optimized beforehand, we were still able to learn the underlying anisotropic character of the true surface through proper selection of training points. Note that there was no need to implement the accuracy measures introduced in Section~\ref{sec:accuracy measure}, since in 2-D one can easily check the quality of the approximation visually.
%%%%%%%%%%%%%%%%%%%%%%%%%%%%%%%% 

%%%%%%%%%%%%%%%%%%%%%%%%%%%%%%%% 

The warped, stretched structure of the 2-D posterior distribution is a signature of the fact that the data provided by the 6 experiments does not provide sufficient amount of information to pinpoint the underlying true reaction parameters. This raises the interesting question of whether the posterior probability distribution would be better constrained, if the experiments carried more information about the parameters. To this end, we repeated the 2-D parameter inference exercise above using additional data provided by experiment 7 in Table~\ref{tab:3node_exps}.  
%%%%%%%%%%%%%%%%%%%%%%%%%%%%%%%% 
\begin{figure}[h!]
\centering
\includegraphics[scale=0.5]{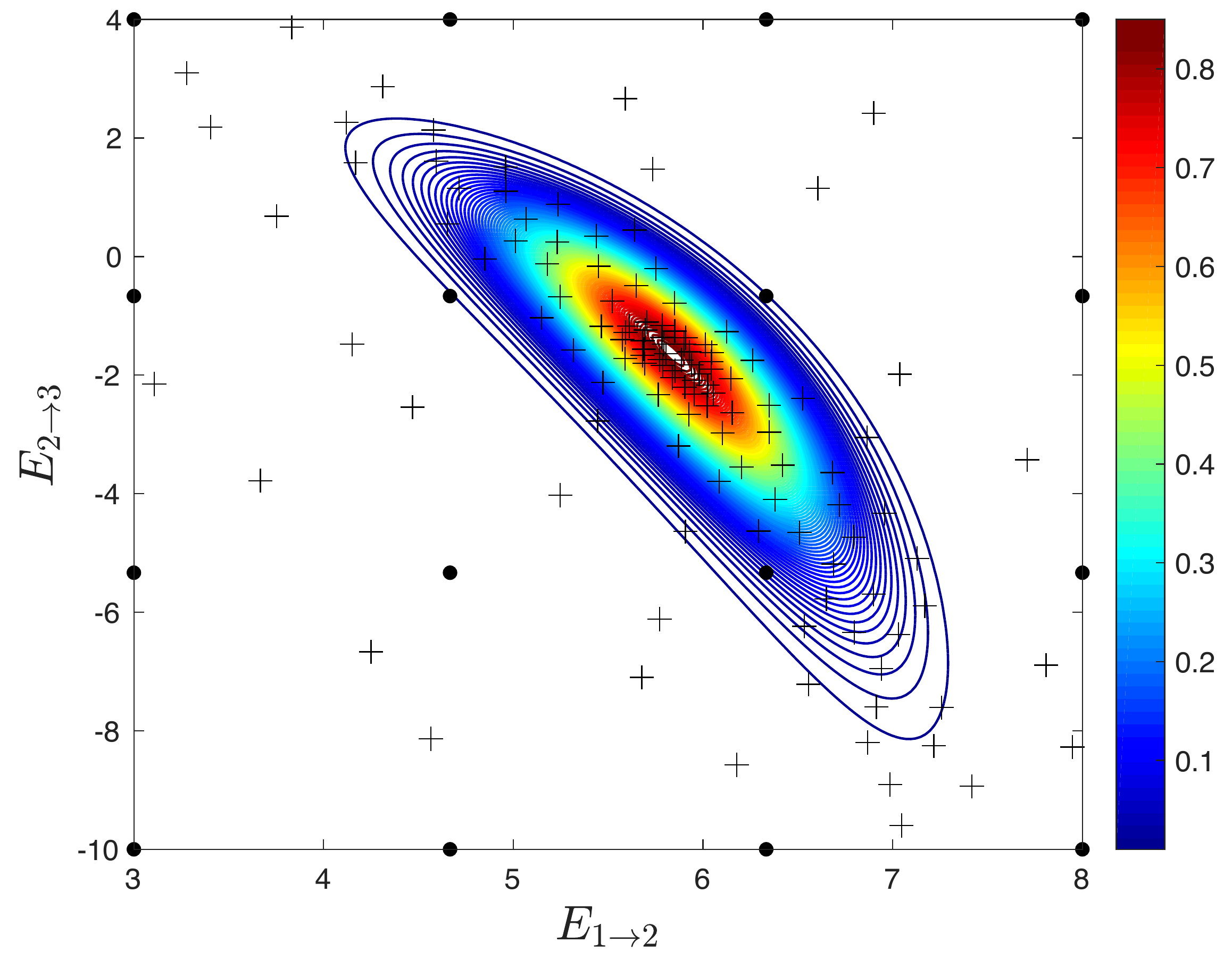}
\caption{Unnormalized true 2-D posterior distribution of $ \Vector \theta = \{ E_{1 \rightarrow 2}, E_{2 \rightarrow 3} \}$ for the 3-node network problem, given a Gaussian prior and observed data from 7 experiments. The black dots represent the initial training data points used for initializing the GP surrogate surface, and the black crosses represent the additional training points selected using the search algorithm.}
\label{fig:truedist-7exp}
\end{figure}

\begin{figure}[h!]
  \begin{subfigure}[t]{0.5\textwidth}
  \includegraphics[scale=0.35]{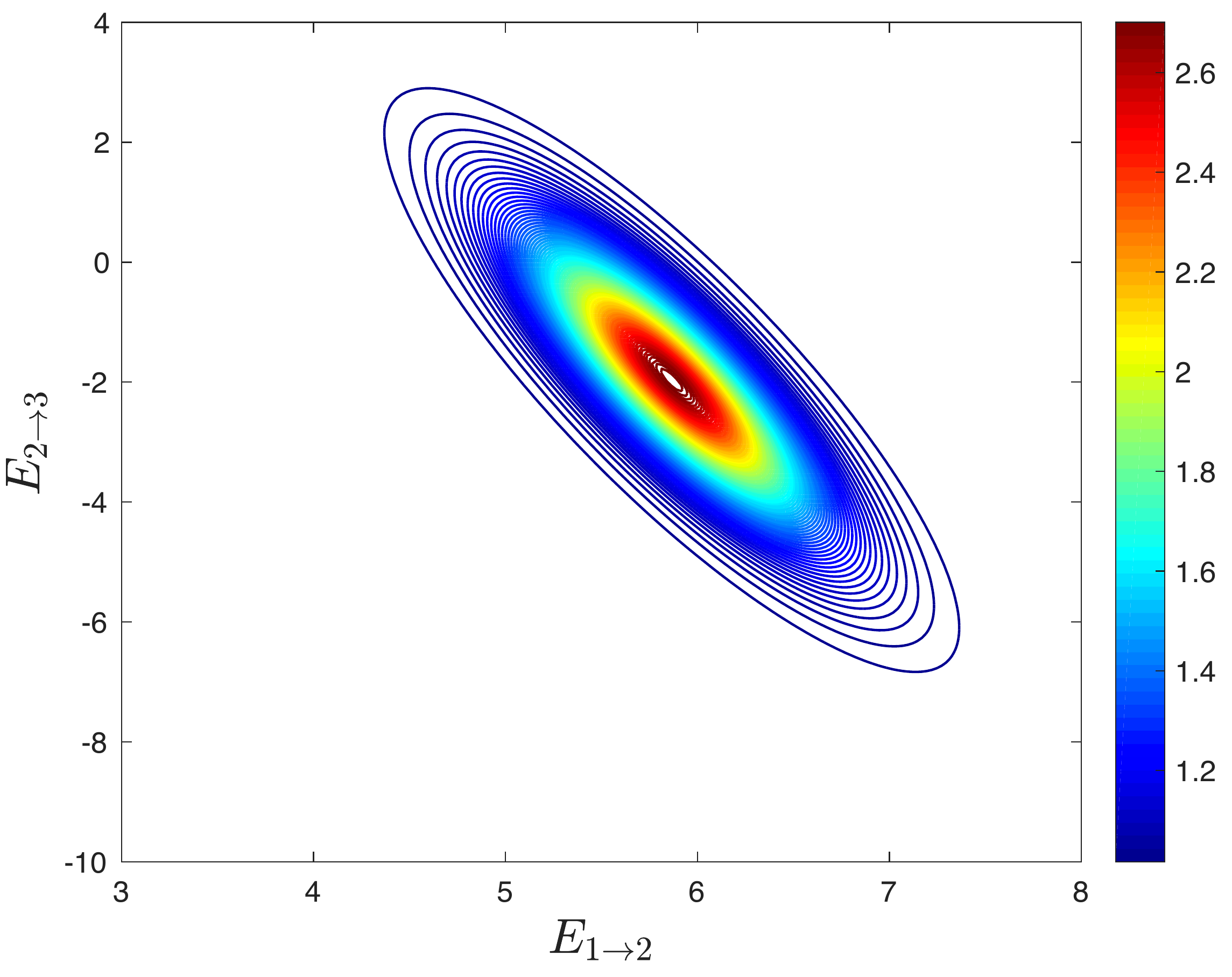}
  \caption{}
  \label{fig:Gauss-7exp}
  \end{subfigure}
  \begin{subfigure}[t]{0.5\textwidth}
  \includegraphics[scale=0.35]{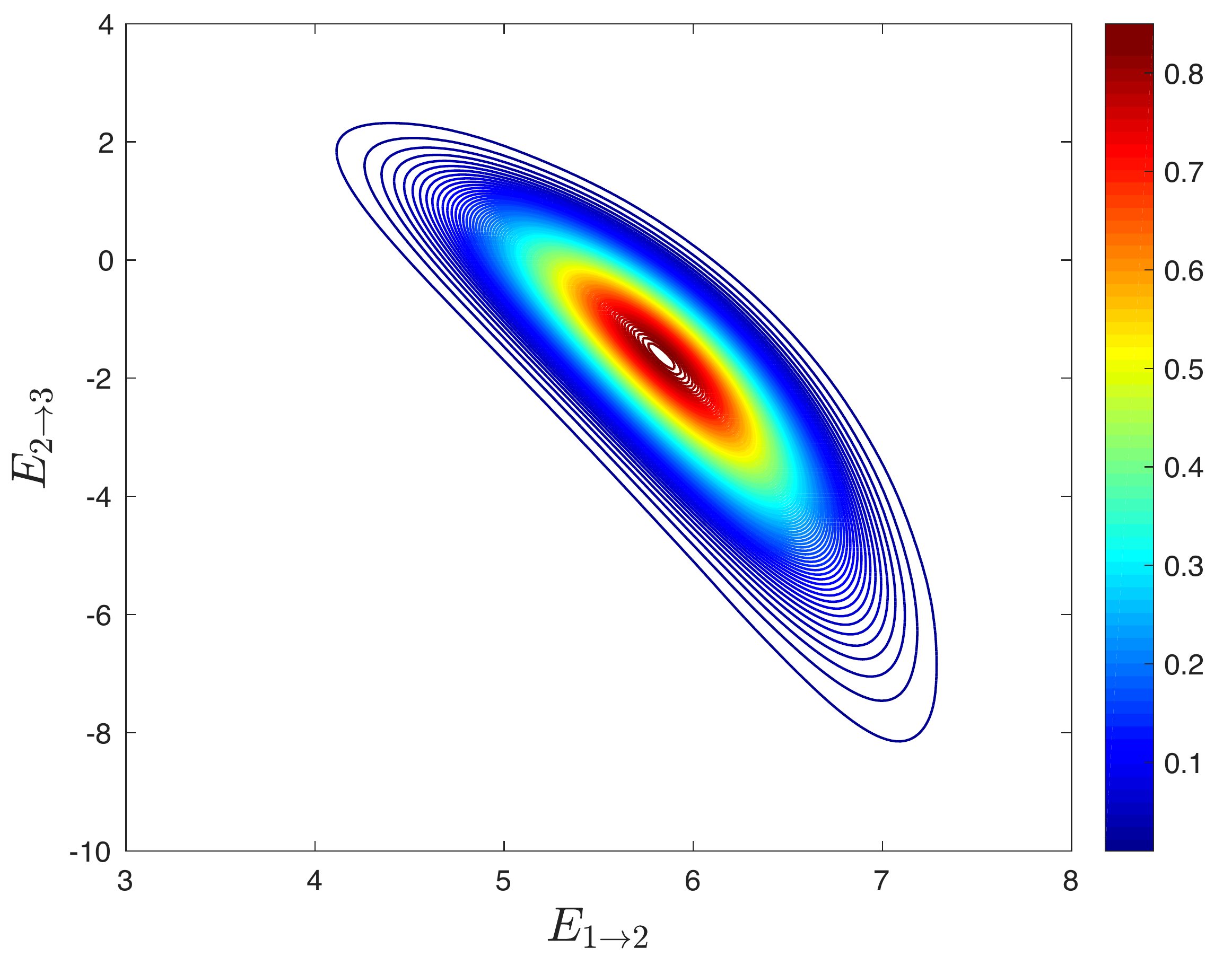}
  \caption{}
  \label{fig:GP200-7exp}
  \end{subfigure}
\caption{ (a) Unnormalized Gaussian prior distribution, $\pi_\text{prior}(\Vector \theta)$ (b) Surrogate posterior distribution, $\widetilde{\pi}_{post}(\Vector \theta)$, constructed using the 216 training points marked in Fig.~(\ref{fig:truedist-7exp}). }
\label{fig:surrogate-7exp}
\end{figure}
%%%%%%%%%%%%%%%%%%%%%%%%%%%%%%%% 
Fig.~(\ref{fig:truedist-7exp}) shows the unnormalized true 2-D posterior distribution, again evaluated on a $200 \times 200$ grid, along with the 216 data input points used to train the GP. The Gaussian prior and the resulting GP-augmented surrogate surface are shown in Fig.~(\ref{fig:surrogate-7exp}). As expected, with the addition of more informative experiments, the posterior is much more constrained. However, the mode of the posterior distribution remains to be a little bit off from the true value of $\Vector \theta = \{ E_{1 \rightarrow 2} = 5, E_{2 \rightarrow 3} = 1 \}$. This is due to the noise in the data and the nonlinear manner by which the activation energy parameter enters the model, which consequently makes the output steeply sensitive to $E$ and thus prohibits exact identification of the truth. Another thing to notice is that the posterior still exhibits a little bit of a curvature, which the surrogate surface manages to capture perfectly.

\subsubsection{6-D case}
\label{sec:6D model}

Having demonstrated that our sequentially trained GP-augmented surrogate surface is capable of successfully reconstructing non-Gaussian 2-D posterior probability distributions, we move on to testing the algorithm on the full 6-D scenario, where we assume that all of the reaction rate parameters are uncertain. In this case, we have $\Vector \theta = \{ A_{1 \rightarrow 2} \, , E_{1 \rightarrow 2} \, , A_{2 \rightarrow 3} \, , E_{2 \rightarrow 3} \, , A_{1 \rightarrow 3} \, , E_{1 \rightarrow 3} \}$. We rely on the same set of noisy data from the 7 experiments shown in Table~\ref{tab:3node_exps} with the same noise level of $\sigma = 0.1$. We use observations from all 7 experiments for the inference this time, since otherwise our problem becomes ill-posed and would require remedies that are beyond the scope of this study.  

\begin{figure}[h!]
  \begin{subfigure}[t]{0.5\textwidth}
  \includegraphics[scale=0.35]{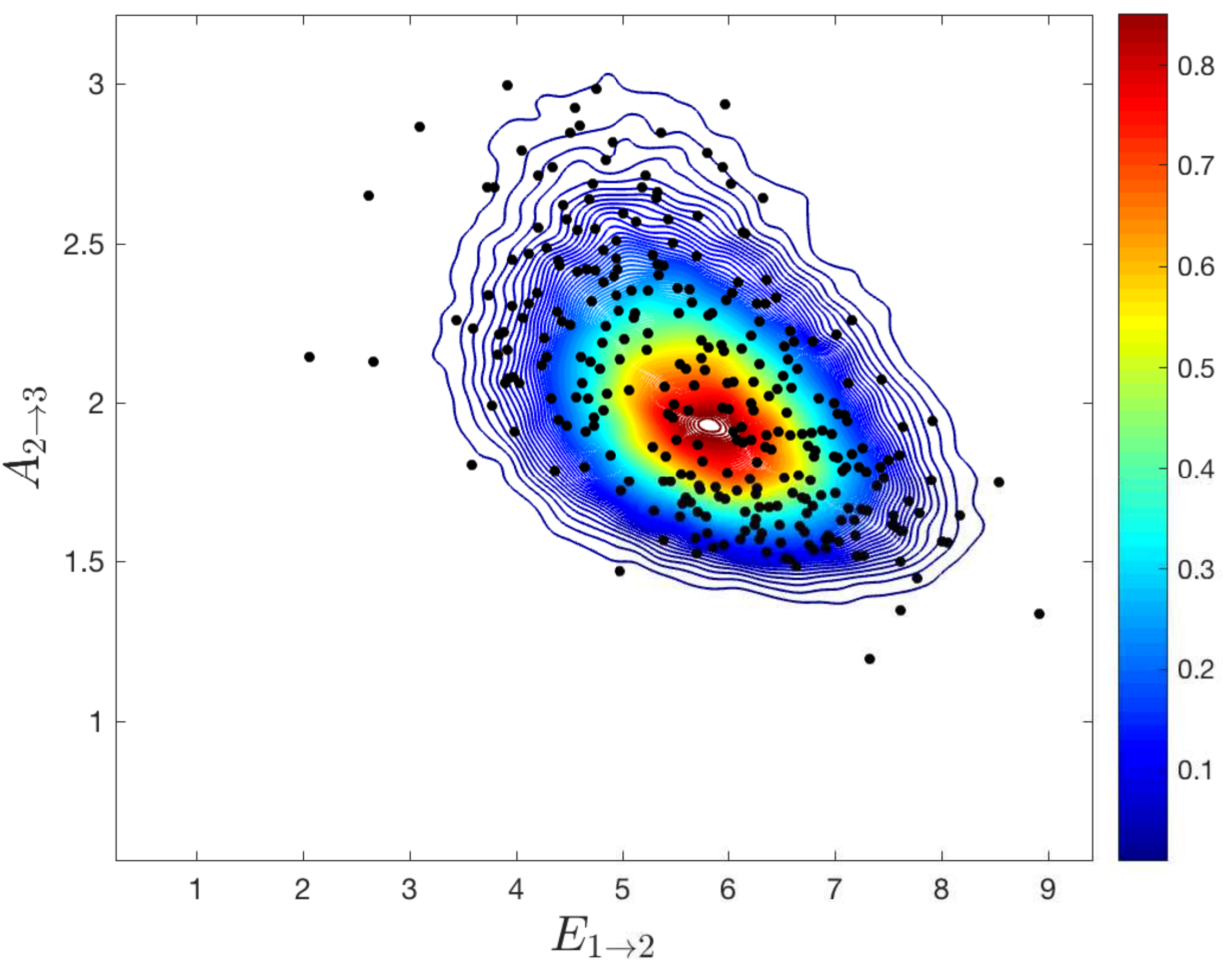}
  \caption{}
  \label{fig:6D-true-1}
  \end{subfigure}
  \begin{subfigure}[t]{0.5\textwidth}
  \includegraphics[scale=0.35]{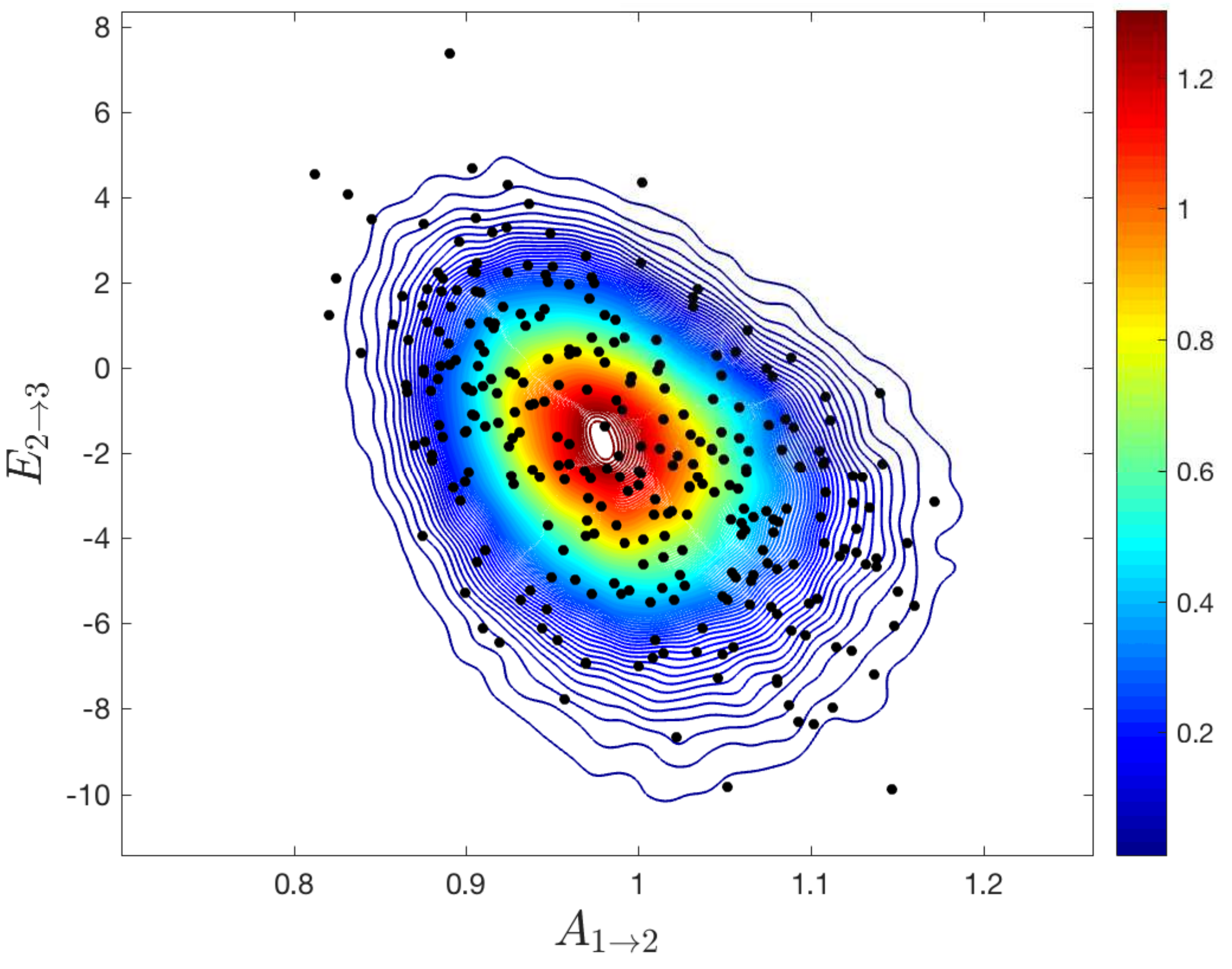}
  \caption{}
  \label{fig:6D-true-2}
  \end{subfigure}
  
  \begin{subfigure}[t]{0.5\textwidth}
  \includegraphics[scale=0.35]{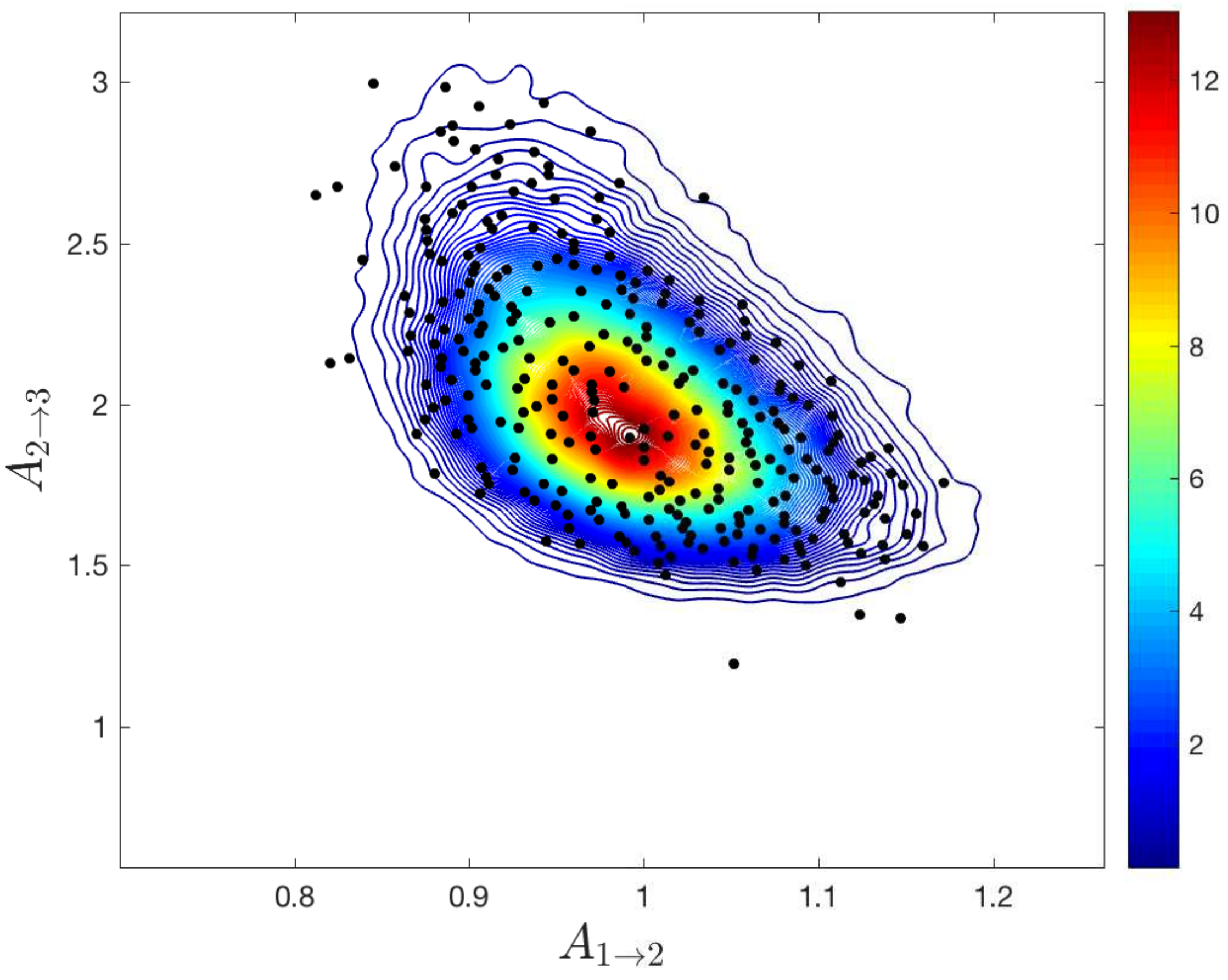}
  \caption{}
  \label{fig:6D-true-3}
  \end{subfigure}
  \begin{subfigure}[t]{0.5\textwidth}
  \includegraphics[scale=0.35]{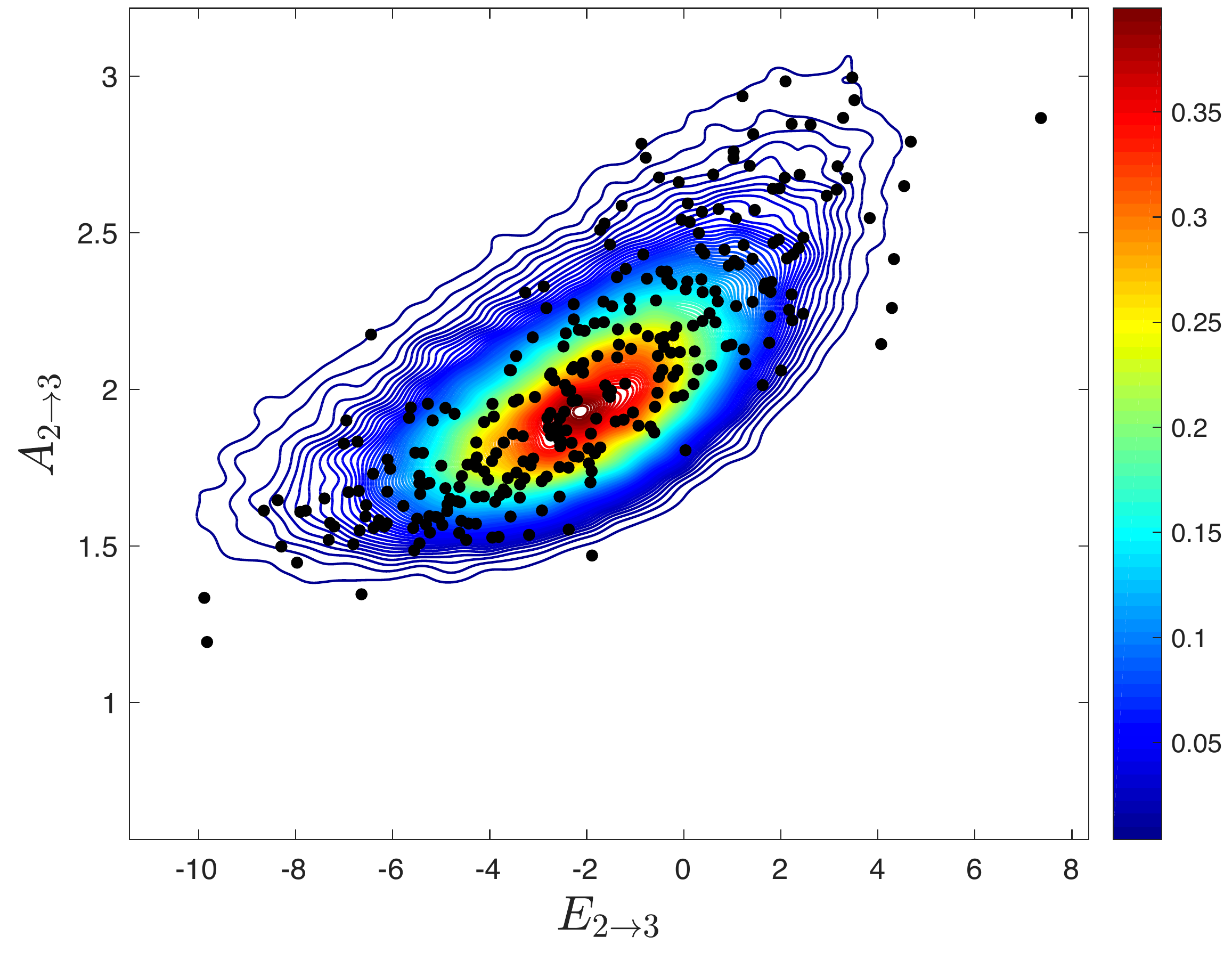}
  \caption{}
  \label{fig:6D-true-4}
  \end{subfigure}
\caption{A subset of 2-D contour projections of the true 6-D posterior probability distribution, $\pi_{\text{post}}(\Vector \theta)$, for the 3-node network problem, given a Gaussian prior and observed data from 7 experiments. The black dots represent projections of the additional GP training points selected using the search algorithm.}
\label{fig:6D-true}
\end{figure}

Fig.~(\ref{fig:6D-true}) shows 2-D contour projections of the true 6-D posterior probability distribution, $\pi_{\text{post}}(\Vector \theta)$, along the dimensions labeled in the figure. The probability contours were obtained by running the Hammer sampler on the true posterior probability surface, collecting the samples after burn-in, projecting them along the given directions, and then estimating the probability using a Gaussian bivariate kernel density estimator (KDE). Rather than dealing with the logarithm of the pre-exponential $A$ parameters in order to impose their positivity constraint, we instead chose to assign a negligibly low probability to negative $A$ values, thus preventing the Hammer sampler from visiting non-admissible regions. By comparing the scales of the axes in Fig.~(\ref{fig:6D-true}), one can notice that the degree of uncertainty in the $E$ parameters is higher than that in the $A$ parameters. This is again due to the fact that the experiments are more informative of the latter than the former.  
%%%%%%%%%%%%%%%%%%%%%%%%%%%%%%%% 
%%%%%%%%%%%%%%%%%%%%%%%%%%%%%%%%

\begin{figure}[p]

  \begin{subfigure}[t]{0.5\textwidth}
  \centering
  \raisebox{-\height}{\includegraphics[scale=0.3]{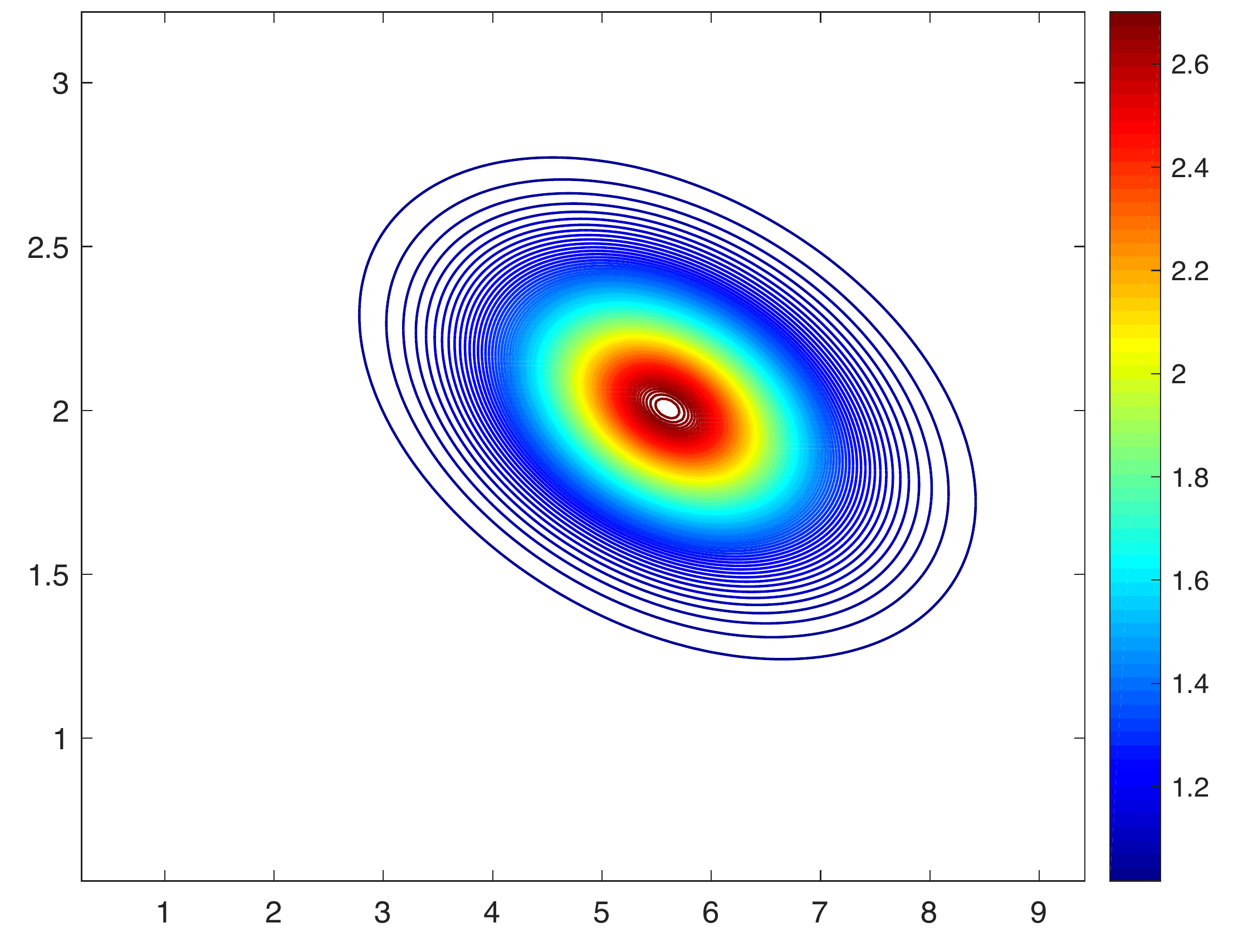}}
  \end{subfigure}
  \begin{subfigure}[t]{0.5\textwidth}
  \centering
  \raisebox{-\height}{\includegraphics[scale=0.3]{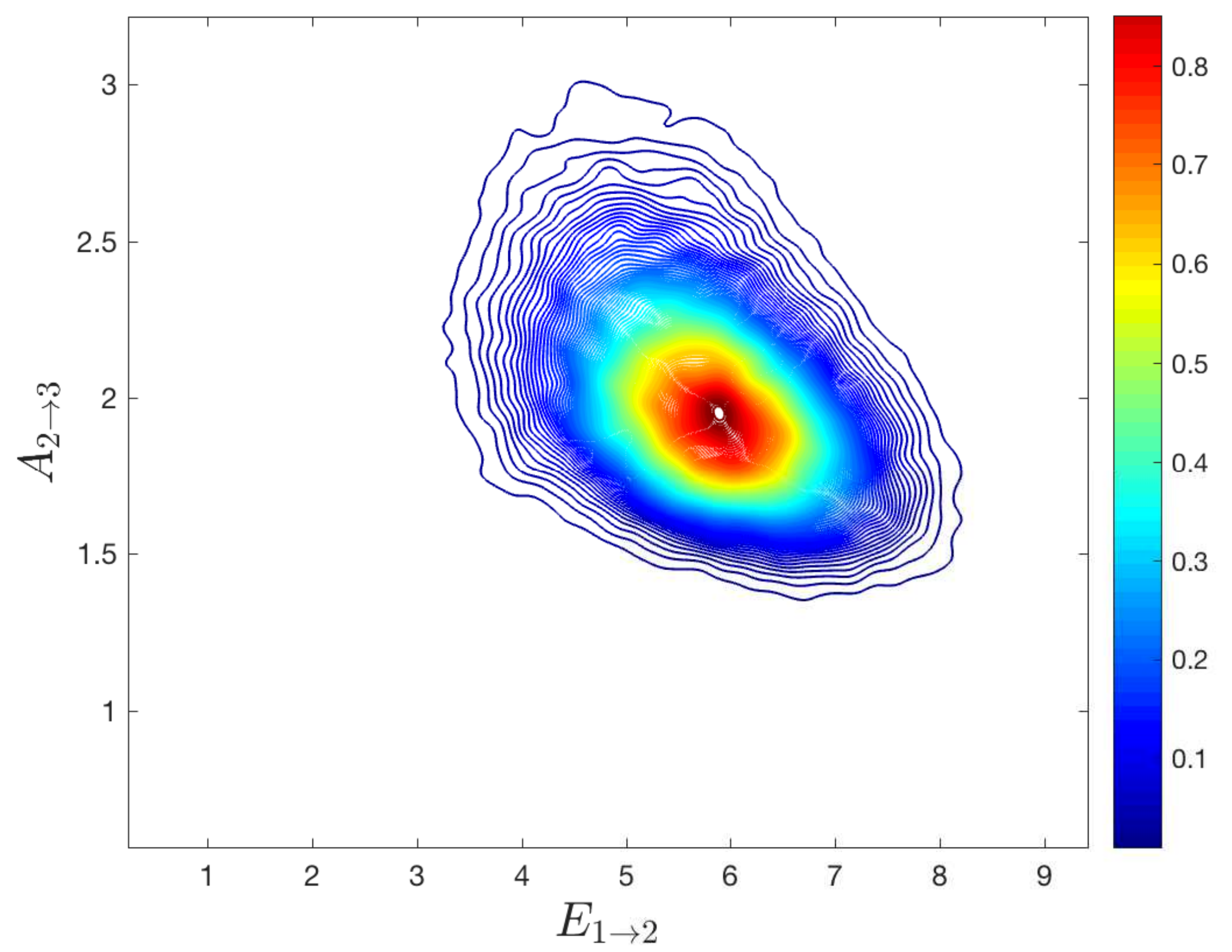}}
  \end{subfigure}
  
  \begin{subfigure}[t]{0.5\textwidth}
  \centering
  \raisebox{-\height}{\includegraphics[scale=0.3]{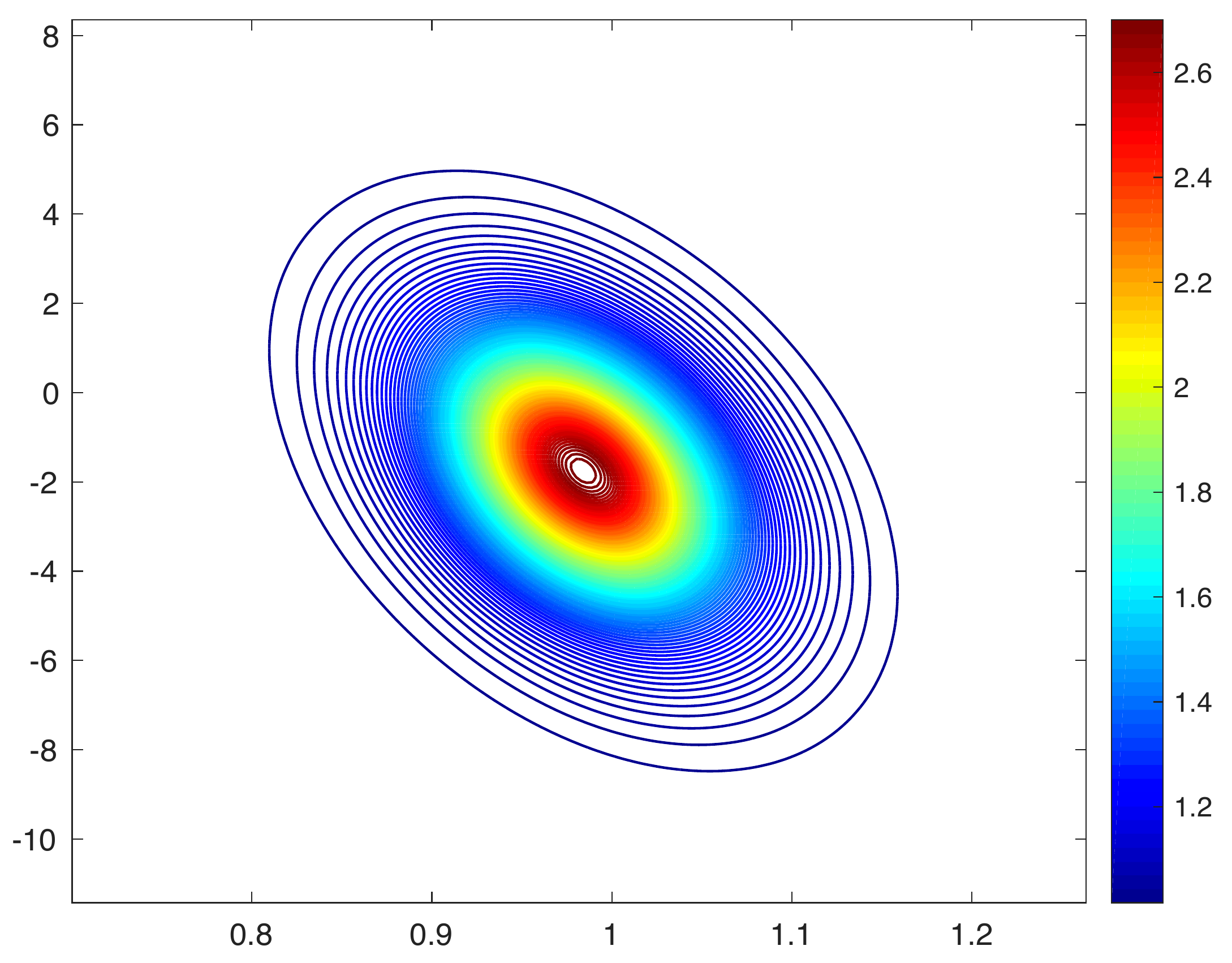}}
  \end{subfigure}
  \begin{subfigure}[t]{0.5\textwidth}
  \centering
  \raisebox{-\height}{\includegraphics[scale=0.3]{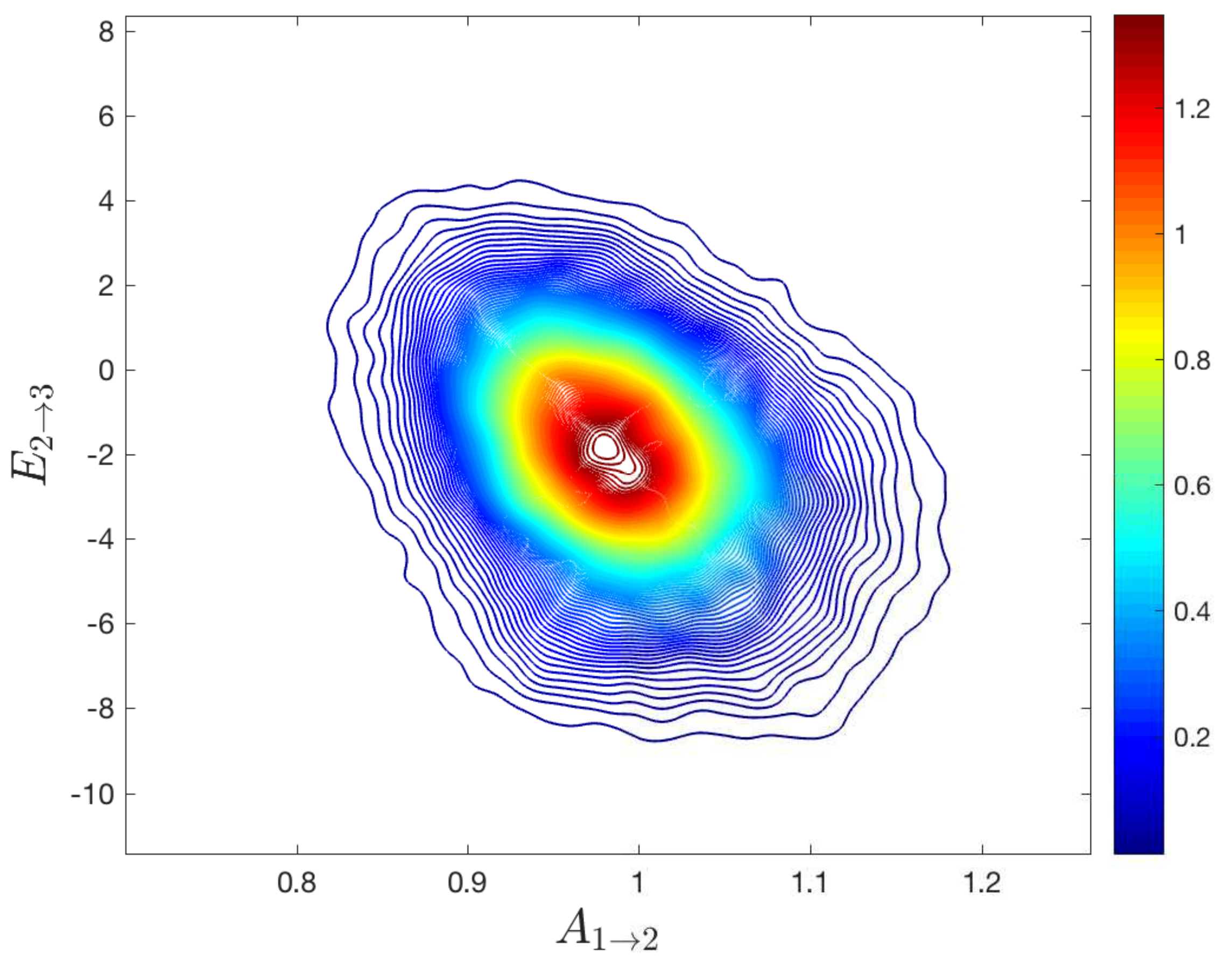}}
  \end{subfigure}  
    
  \begin{subfigure}[t]{0.5\textwidth}
  \centering
  \raisebox{-\height}{\includegraphics[scale=0.3]{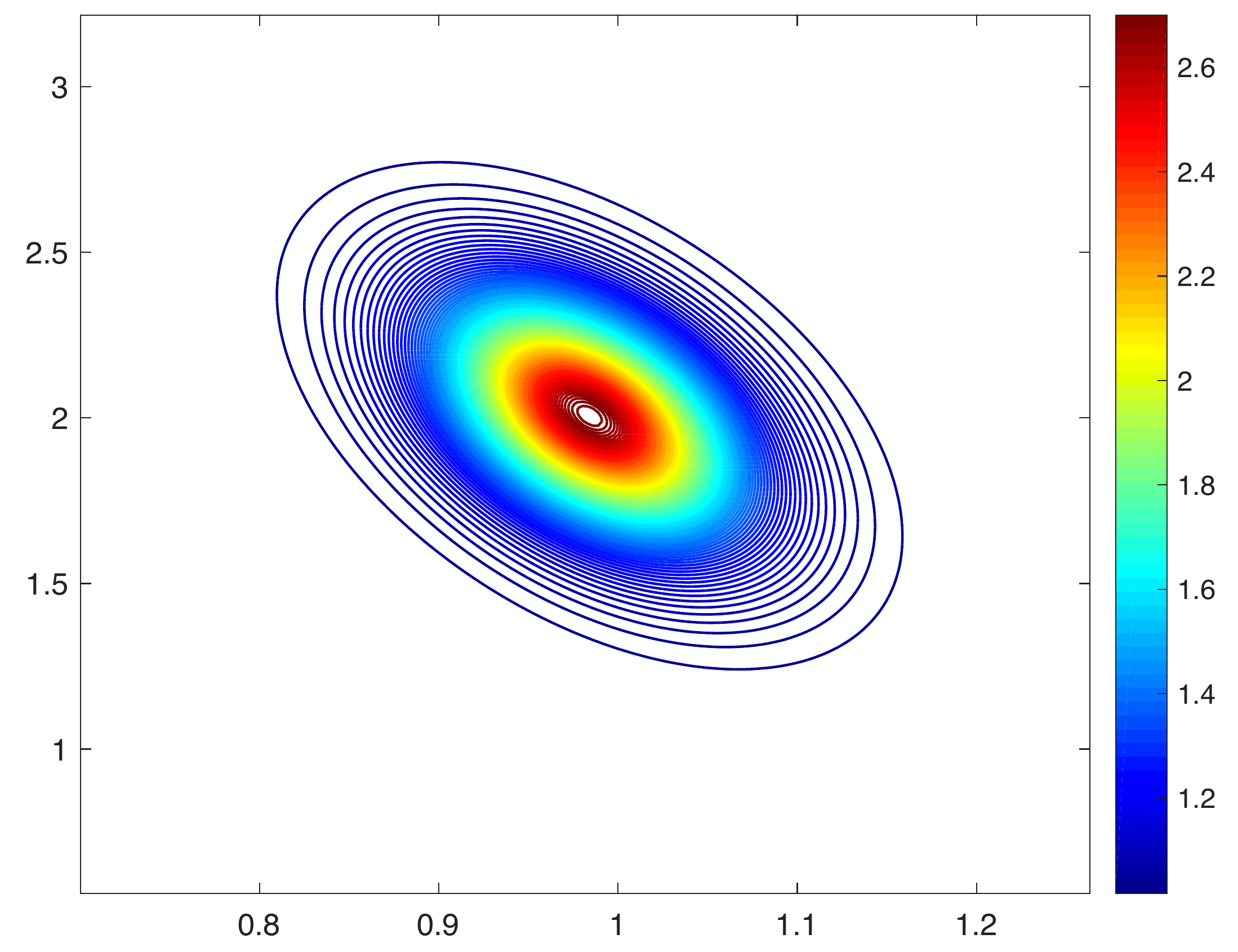}}
  \end{subfigure}
  \begin{subfigure}[t]{0.5\textwidth}
  \centering
  \raisebox{-\height}{\includegraphics[scale=0.3]{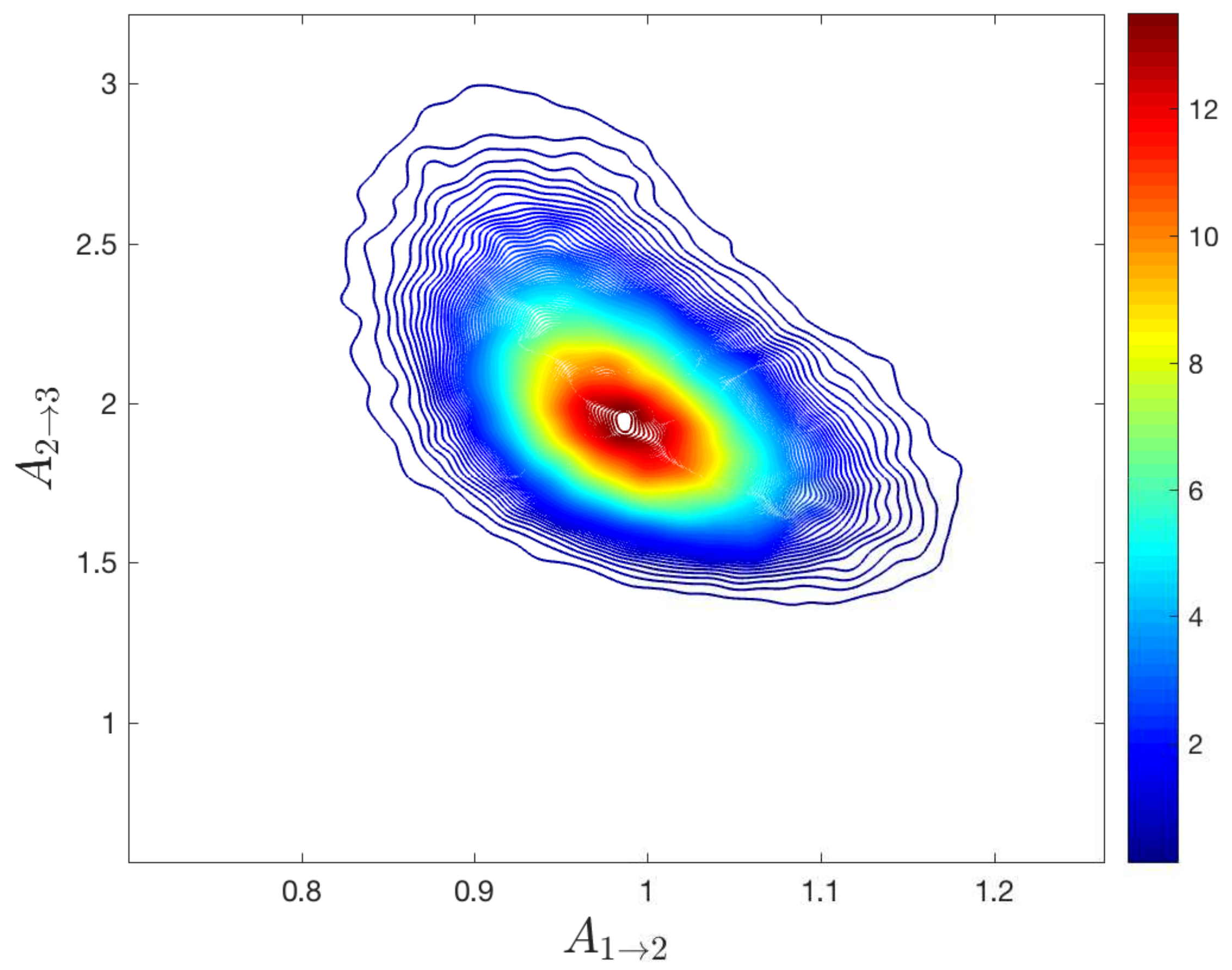}}
  \end{subfigure}  
 
  \begin{subfigure}[t]{0.5\textwidth}
  \centering
  \raisebox{-\height}{\includegraphics[scale=0.3]{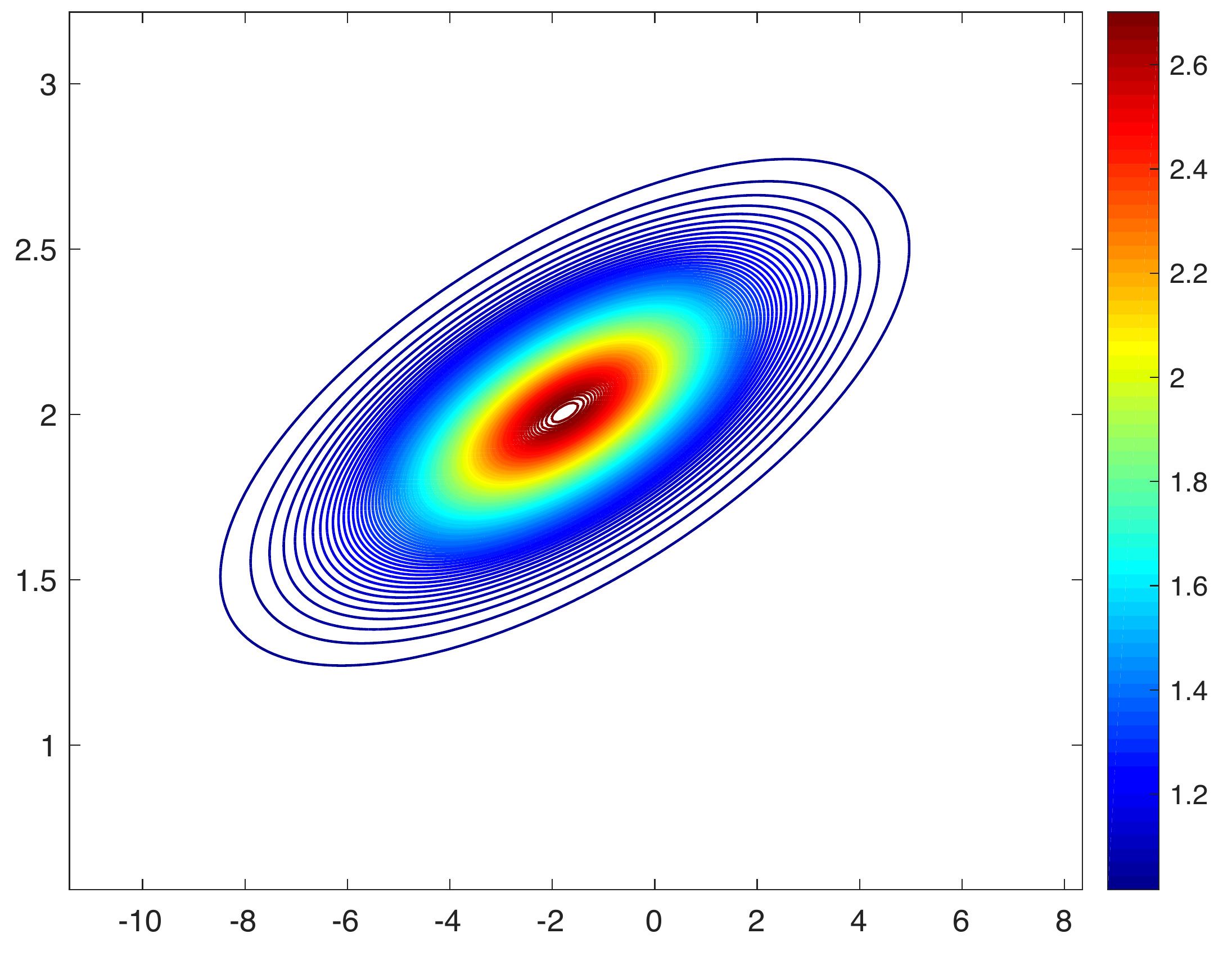}}
  \end{subfigure}
  \begin{subfigure}[t]{0.5\textwidth}
  \centering
  \raisebox{-\height}{\includegraphics[scale=0.3]{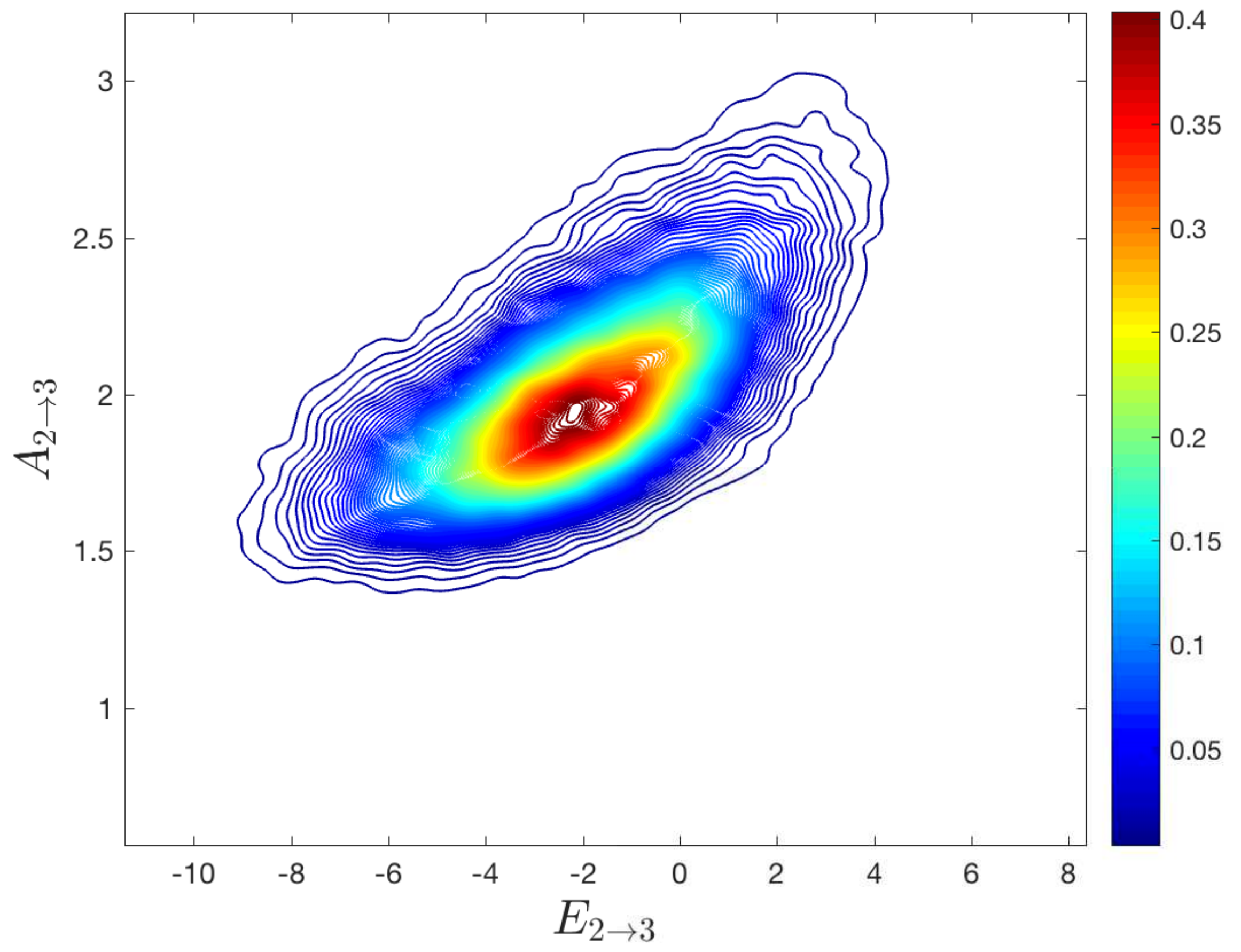}}
  \end{subfigure} 
  
\caption{Left column corresponds to projections of the unnormalized Gaussian prior distribution, $\pi_\text{prior}(\Vector \theta)$, and right column corresponds to projections of the resulting 6-D surrogate posterior distribution, $\widetilde{\pi}_\text{post}(\Vector \theta)$ after the addition of the training points marked in Fig.~(\ref{fig:6D-true}).}
\label{fig:6D-surrogate}
\end{figure}

In order to construct our surrogate surface, we started as before by constructing the Gaussian $\pi_\text{prior}(\Vector \theta)$, and initializing the GP surface with a set of training points. 
The training points for initializing the GP surface were obtained by collecting sample points from a random selection of 10 iterations of the Hammer chain (after burn-in) utilized in constructing $\pi_\text{prior}(\Vector \theta)$. Note that we did not impose the minimum distance constraint, mentioned in Section~\ref{sec:optimal point}, on this initial set of training points. The resulting surrogate surface was then sequentially updated with 300 additional observation points chosen according to our search algorithm (with the minimum distance constraint imposed this time). 2-D projections of the additional 300 training points selected by the search algorithm are shown overlaid on the corresponding projections of the true posterior distribution in Fig.~(\ref{fig:6D-true}). For clarity, the initial GP training points have not been included in the plots. The initial Gaussian prior and the final surrogate distributions are shown in the left and right columns, respectively, of Fig.~(\ref{fig:6D-surrogate}). The rows in  Fig.~(\ref{fig:6D-surrogate}) correspond to projections of the prior and surrogate distributions along the same directions as those shown in Figs.~(\ref{fig:6D-true-1})--(\ref{fig:6D-true-4}). The probability contours of the surrogate posterior distribution were obtained via KDE in the same manner as those in Fig.~(\ref{fig:6D-true}), however, the contours of the initial Gaussian prior were obtained by numerically evaluating the 2-D marginal distributions on a $200 \times 200$ grid in the range shown. As can be seen from the figures, the surrogate surface successfully captures all of the non-linearities present in the true posterior distribution, which the initial Gaussian prior approximation fails to represent. 
%%%%%%%%%%%%%%%%%%%%%%%%%%%%%%%% 
%%%%%%%%%%%%%%%%%%%%%%%%%%%%%%%%

\begin{figure}[h!]
  \begin{subfigure}[t]{0.5\textwidth}
  \centering
  \includegraphics[scale=0.35]{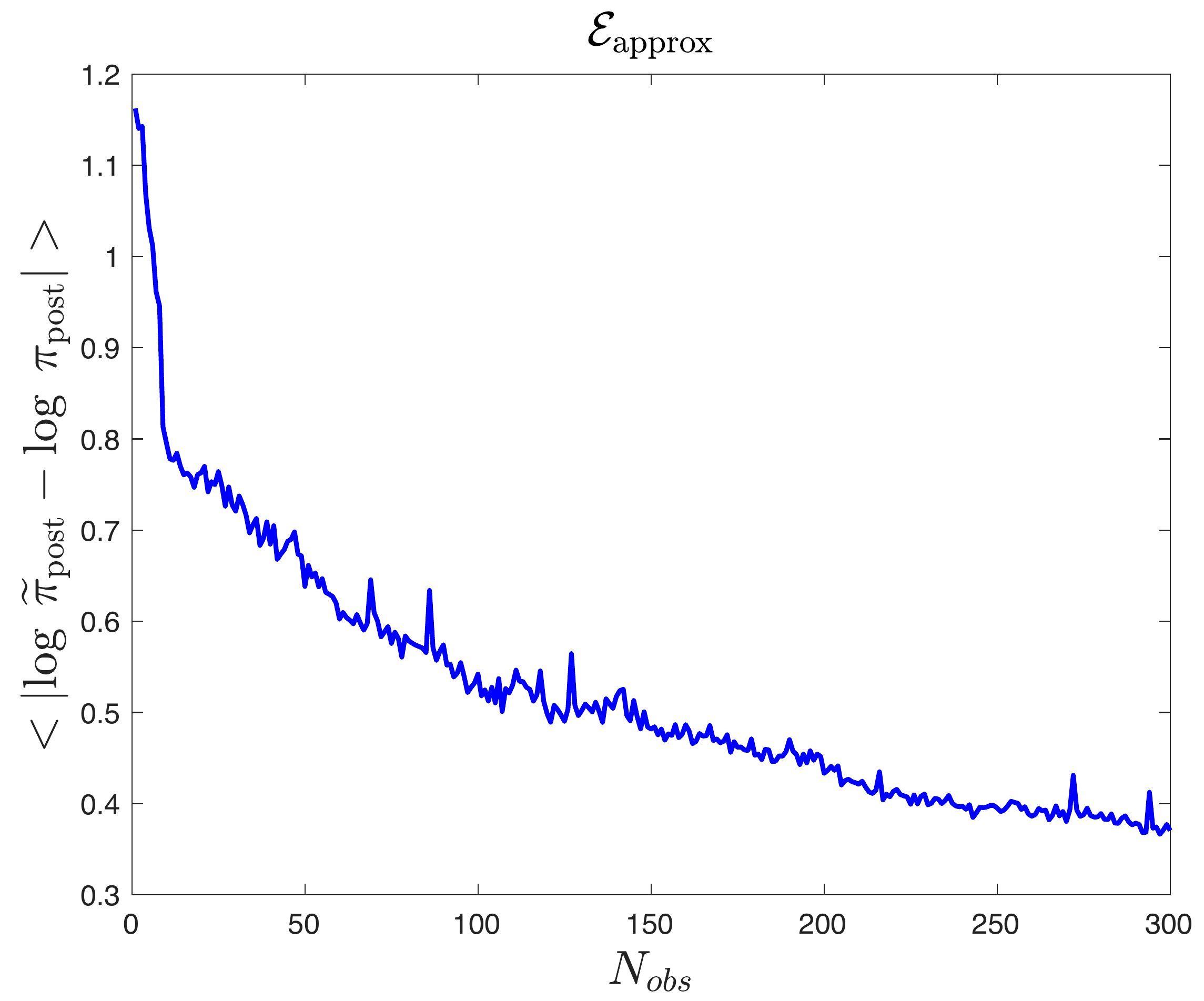}
  \caption{}
  \label{fig:6D-Egp}
  \end{subfigure}
  \begin{subfigure}[t]{0.5\textwidth}
  \centering
  \includegraphics[scale=0.35]{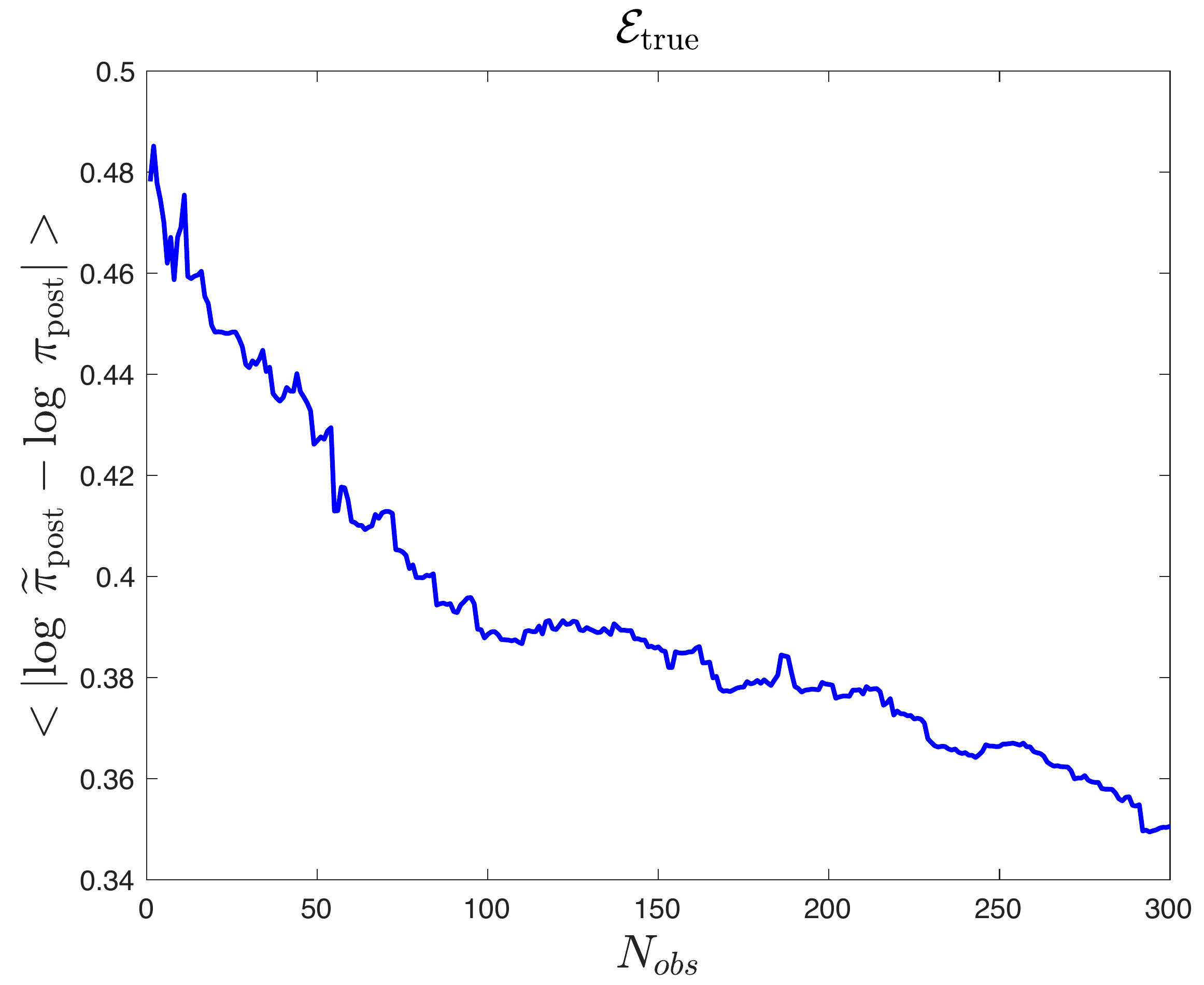}
  \caption{}
  \label{fig:6D-Etrue}
  \end{subfigure}
  \begin{subfigure}[t]{\textwidth}
  \centering
  \includegraphics[scale=0.35]{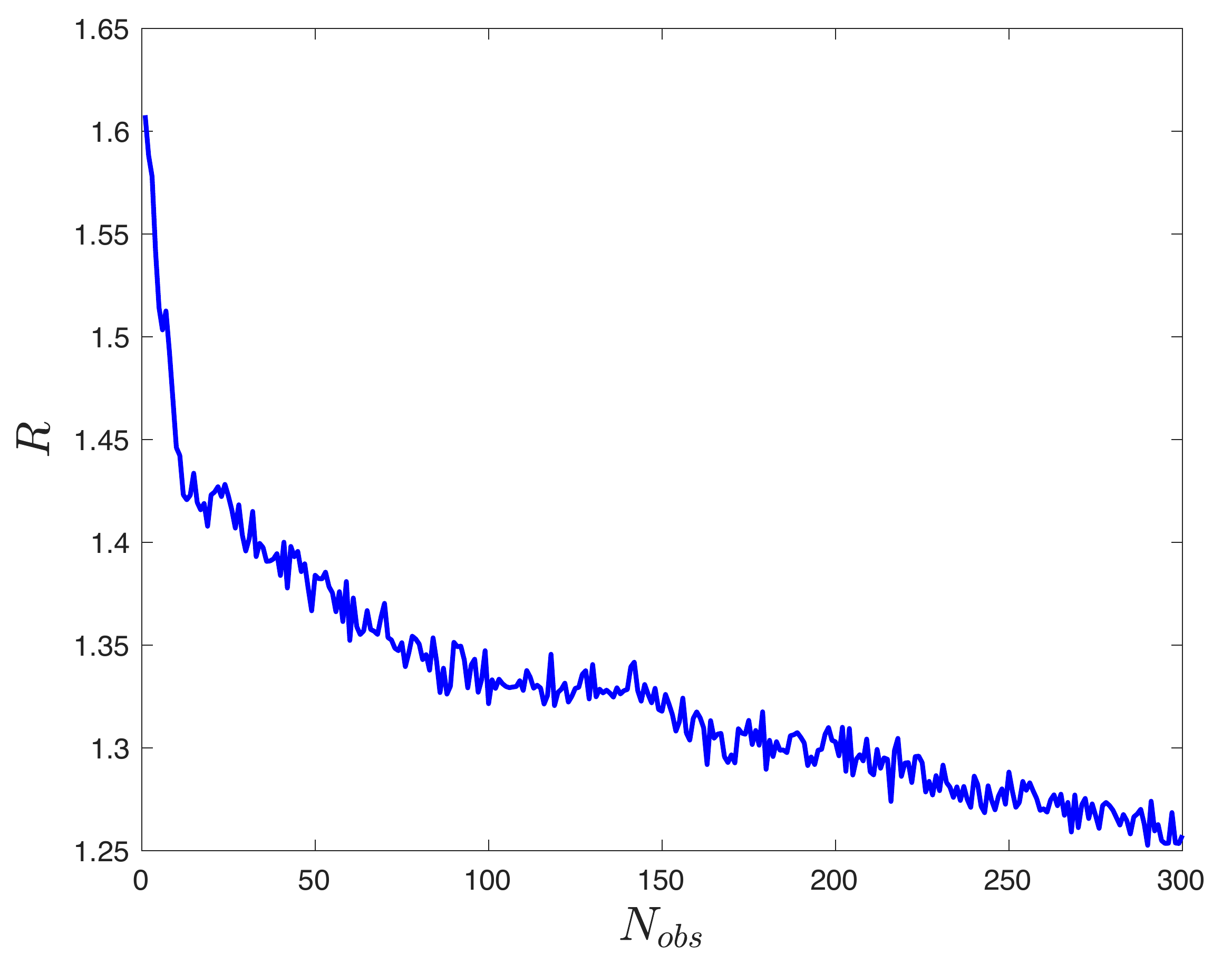}
  \caption{}
  \label{fig:6D-Rmeasure}
  \end{subfigure}
\caption{Evolution of the emperical average absolute errors and the $R$ measure of the 6-D surrogate surface as training points are added sequentially to the GP surface.}
\label{fig:6D-error}
\end{figure}

To better quantify the accuracy of the surrogate approximation, especially since it is not easy to directly visualize a 6-D surface, and to be able to judge the efficiency of our sequential training algorithm, we resort to the accuracy measures given by Eqs.~(\ref{eq:abs_error})--(\ref{eq:Rmeasure}). Figs.~(\ref{fig:6D-Egp})--(\ref{fig:6D-Rmeasure}) show the evolution of $\mathcal E_\text{approx}$, $\mathcal E_\text{true}$, and the $R$ measure, respectively, as we sequentially add training points to the GP-augmented surface. Note that $N_{obs} = 0$ corresponds to the GP-augmented surface that has already been trained with the initializing set of GP training points. 
As can be seen from Figs.~(\ref{fig:6D-error}a-b), the initial values of $\mathcal E_\text{approx}$ and $\mathcal E_\text{true}$ are not appreciably large, indicating that the initial GP-augmented surface is not vastly different from the true posterior surface. Notice also that the initial $\mathcal E_\text{approx} > \mathcal E_\text{true}$, which means that the initial GP-augmented surface is more accurate near the mode of the posterior distribution than farther away. This is expected, since (in most cases) a Gaussian approximation near the mode of the true posterior distribution should locally be fairly accurate. Considering also that the covariance kernel correlation length-scale is not too large, the initial GP-augmented surface starts out fairly lumpy as it adjusts its mean predictions with the addition of training points. So in computing $\mathcal E_\text{approx}$, more sample points initially come from regions where the surrogate surface is still not very accurate and where the true probability is low. On the other hand, the sample points used to compute $\mathcal E_\text{true}$ are always more concentrated near the mode of the true posterior distribution, where the surrogate surface is initially relatively more accurate. This underscores the importance of relying on both measures to properly judge the accuracy of the surrogate surface.
Another thing to notice is that both error values exhibit an overall monotonic decrease as more observation points are added, which confirms that the search algorithm is selecting new training points that are indeed informative. Moreover, the decrease in error appears to be steepest at the beginning, and becomes more gradual later on. This too confirms that the algorithm is seeking training points that are maximally informative in a global sense, as desired.
%%%%%%%%%%%%%%%%%%%%%%%%%%%%%%%%
%%%%%%%%%%%%%%%%%%%%%%%%%%%%%%%%

Observing now Fig.~(\ref{fig:6D-Rmeasure}), we can notice (i) that the initial R measure ($R = 1.61$) is only slightly larger than $R = 1$, which indicates that the initial surrogate approximation, while not exact, is still appreciably close to the true posterior surface, (ii) that the R-measure exhibits an overall monotonic decreasing trend as more training points are added to the GP surface, and (iii) that the steepest decrease occurs when the first few training points are added, while the rate of decrease becomes more gradual later on. These observations are consistent and lend further support to our earlier remarks regarding the absolute error trends.
\clearpage

\subsection{6-node network model}
\label{sec:6-node network}

\begin{figure}[h!]
\centering
\includegraphics[scale=0.7]{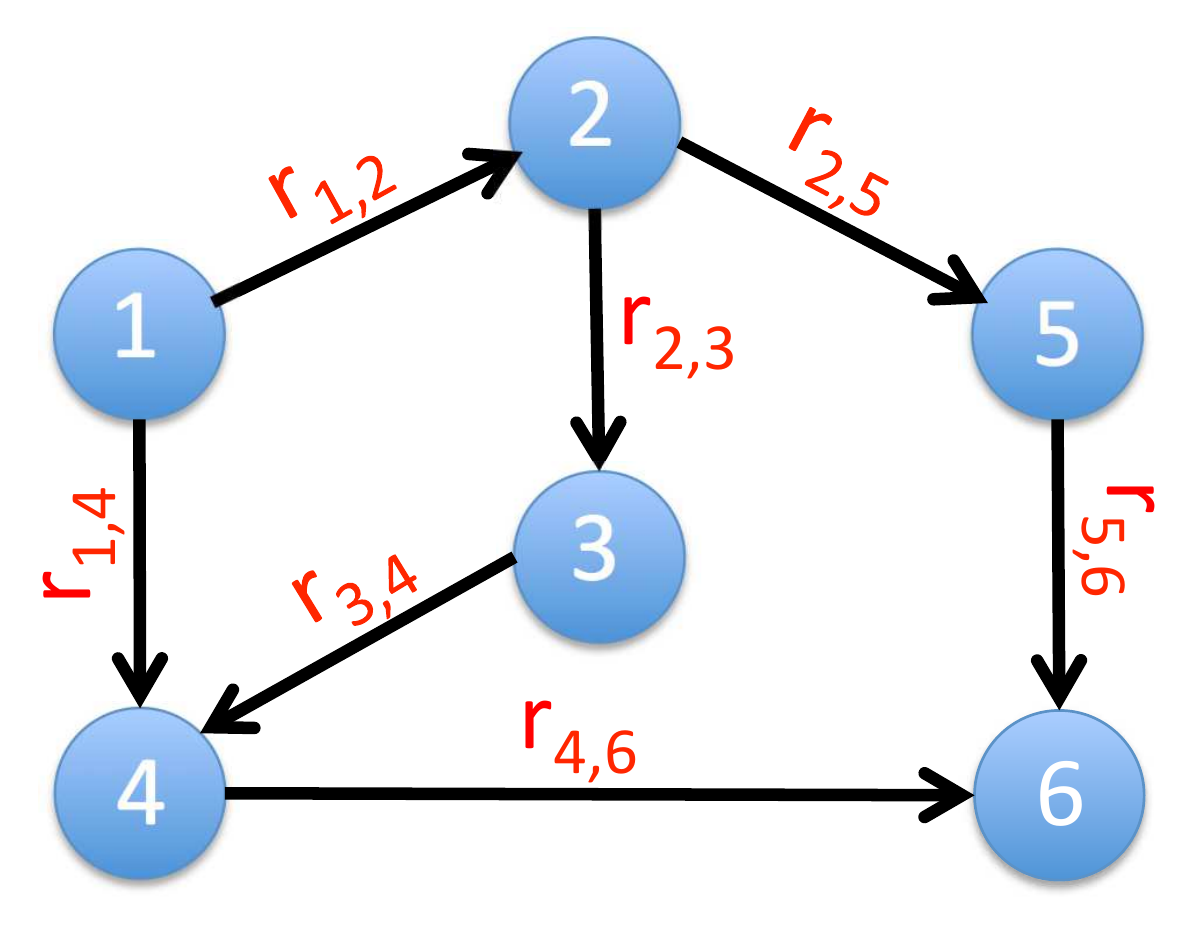}
\caption{Schematic of the 6-node network model.}
\label{fig:6node_schematic}
\end{figure}

In this section, we consider a slightly more complicated 6-node network model as illustrated in the schematic in Fig.~(\ref{fig:6node_schematic}). With the exception of being composed of more reactive nodes, this 6-node network model is governed by the same assumptions and reaction mechanisms as those of the 3-node network model. In the current model, however, we assume that only the pre-exponential, $A$, parameters are uncertain, which makes our inference problem 7-D with $\Vector \theta = \{ A_{1 \rightarrow 2} \, , A_{1 \rightarrow 4} \, , A_{2 \rightarrow 3} \, , A_{2 \rightarrow 5} \, , A_{3 \rightarrow 4} \, , A_{4 \rightarrow 6} \, , A_{5 \rightarrow 6} \}$. 

%%%%%%%%%%%%%%%%%%%%%%%%%%%%%%%% 

%%%%%%%%%%%%%%%%%%%%%%%%%%%%%%%% 

For inferring the uncertain $A$ parameters, we rely, as before, on observations from a set of 20 synthetic noisy experiments with a noise level of $\sigma = 0.6$. We chose a higher noise level this time, in order to increase the degree of non-Gaussianity of the posterior distribution and make the problem slightly more challenging. The true network reaction rate parameters are shown in Table~\ref{tab:6node_params}, and Table~\ref{tab:6node_exps} lists the conditions for each experiment. Note that each of the 10 experiments shown in Table~\ref{tab:6node_exps} was carried out with $\beta = 0.01$. The remaining 10 experiments (not listed) were carried out at the same conditions as those listed in the table, but with $\beta = 0.1$. 
%%%%%%%%%%%%%%%%%%%%%%%%%%%%%%%% 
\begin{table}[h]
\begin{center}
\begin{tabular} {| *{8}{c} |}
\hline
Nodes & (1,2) & (1,4) & (2,3) & (2,5) & (3,4) & (4,6) & (5,6) \\ \hline
$E$ & 5 & 2 & 2 & 4 & 4 & 3 & 2 \\ \hline
$A$ & 7 & 2 & 3 & 6 & 5 & 4 & 1 \\
\hline
\end{tabular} 
\end{center}
\caption{True parameters of the 6-node network model. The rate of the reaction across each node is given by $r_{i,j;k} = \text{C}_{i,j; k} \, A_{i,j} \, e^{-\beta_k E_{i,j}}$.}
\label{tab:6node_params}
\end{table}

\begin{table}[h]
\begin{center}
\begin{tabular} {| c | *{7}{c} |}
\hline
Experiment $k$ & $\text{C}_{1 \rightarrow 2}$ & $\text{C}_{1 \rightarrow 4}$ & $\text{C}_{2 \rightarrow 3}$ & $\text{C}_{2 \rightarrow 5}$ & $\text{C}_{3 \rightarrow 4}$ & $\text{C}_{4 \rightarrow 6}$ & $\text{C}_{5 \rightarrow 6}$\\ \hline
1 & 10 & 0.1 & 10 & 10 & 10 & 0.1 & 10 \\ \hline
2 & 0.1 & 10 & 10 & 0.1 & 10 & 10 & 0.1 \\ \hline
3 & 0.1 & 10 & 0.1 & 10 & 0.1 & 0.1 & 10 \\ \hline
4 & 10 & 0.1 & 10 & 10 & 10 & 10 & 10 \\ \hline
5 & 10 & 10 & 10 & 10 & 10 & 0.1 & 10 \\ \hline
6 & 10 & 10 & 10 & 10 & 0.1 & 10 & 10 \\ \hline
7 & 10 & 10 & 0.1 & 10 & 10 & 10 & 10 \\ \hline
8 & 10 & 10 & 10 & 0.1 & 10 & 10 & 10 \\ \hline
9 & 10 & 10 & 10 & 10 & 10 & 10 & 0.1 \\ \hline
10 & 0.1 & 10 & 10 & 10 & 10 & 10 & 10 \\ 
\hline
\end{tabular} 
\end{center}
\caption{Experiments used for inferring the uncertain reaction rate parameters in the 6-node network model.
These experiments were carried out with $\beta = 0.01$. The same 10 experiments above were repeated with $\beta = 0.1$, resulting in a total of 20 experiments that were employed in the parameter inference problem.}
\label{tab:6node_exps}
\end{table}
%%%%%%%%%%%%%%%%%%%%%%%%%%%%%%%%

\begin{figure}[h!]
  \begin{subfigure}[t]{0.5\textwidth}
  \includegraphics[scale=0.35]{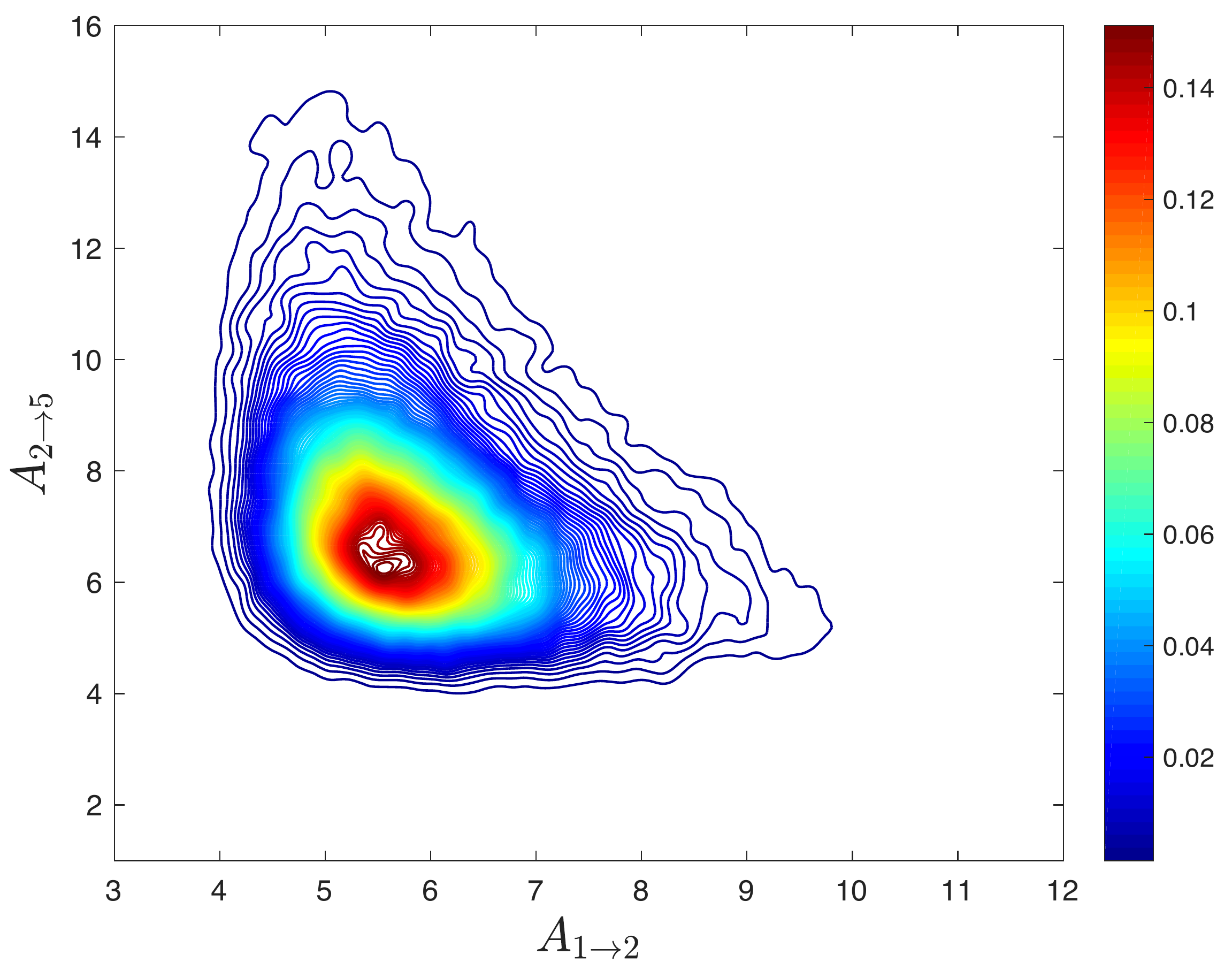}
  \caption{}
  \label{fig:7D-true-1}
  \end{subfigure}
  \begin{subfigure}[t]{0.5\textwidth}
  \includegraphics[scale=0.35]{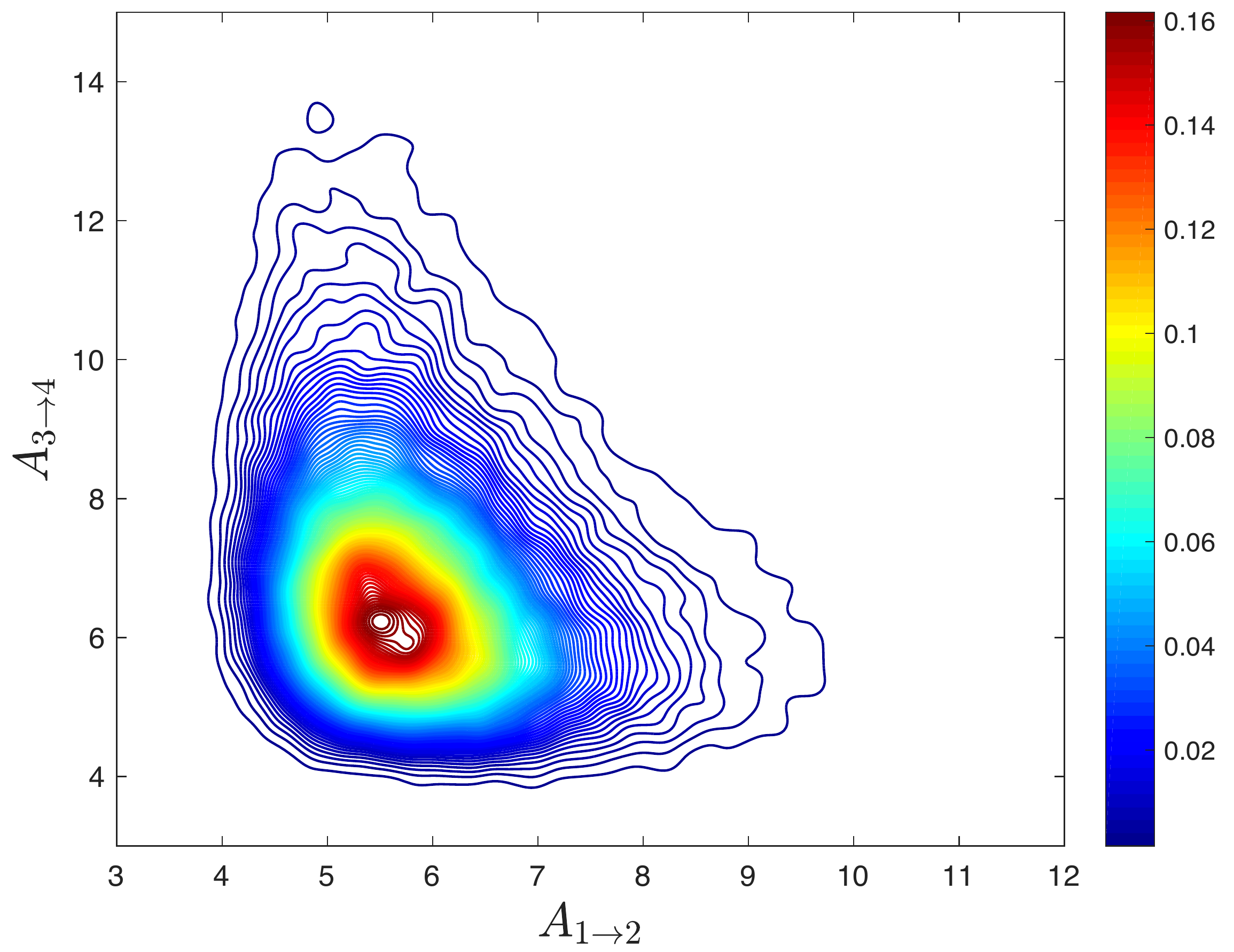}
  \caption{}
  \label{fig:7D-true-2}
  \end{subfigure}
  
  \begin{subfigure}[t]{0.5\textwidth}
  \includegraphics[scale=0.35]{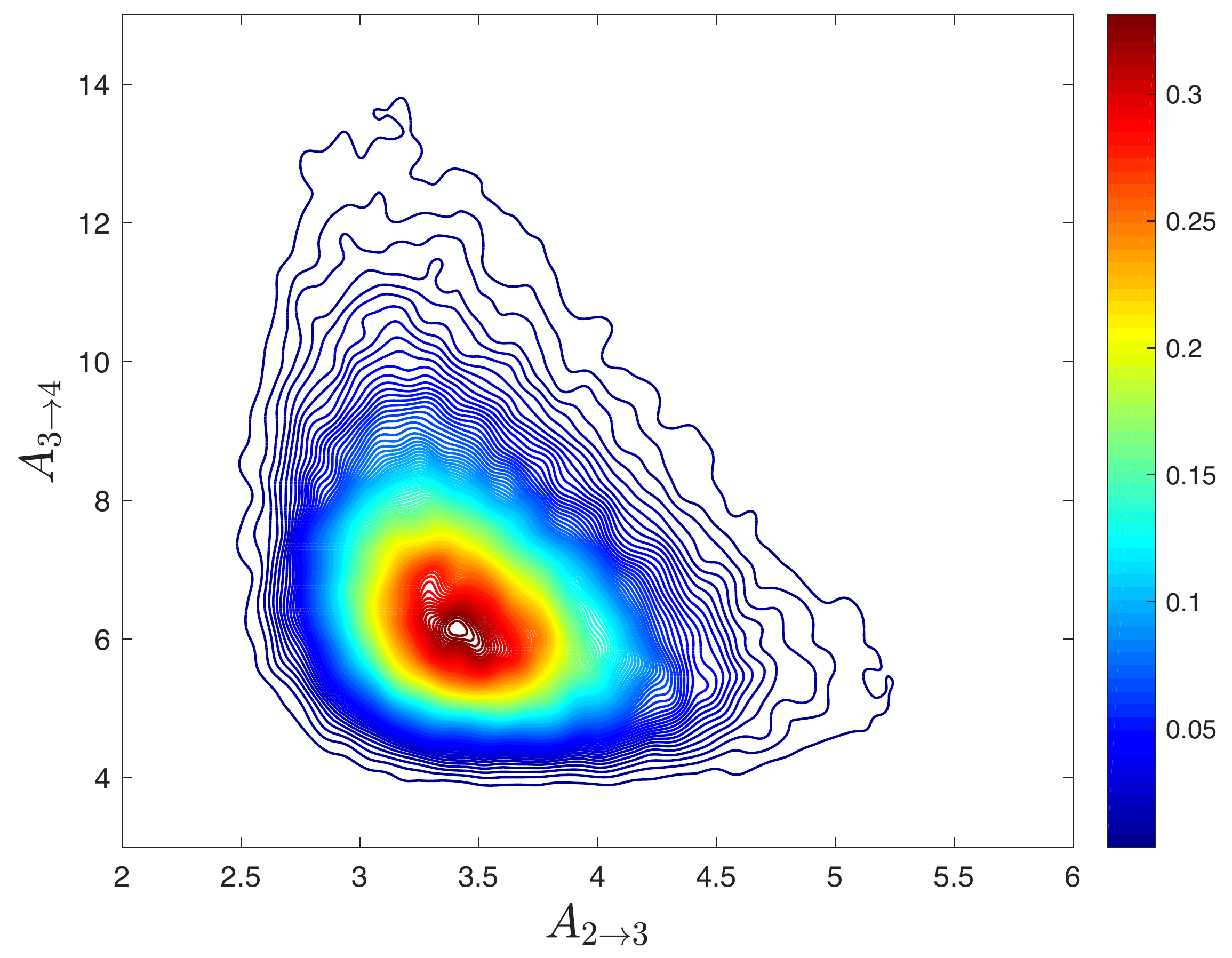}
  \caption{}
  \label{fig:7D-true-3}
  \end{subfigure}
  \begin{subfigure}[t]{0.5\textwidth}
  \includegraphics[scale=0.35]{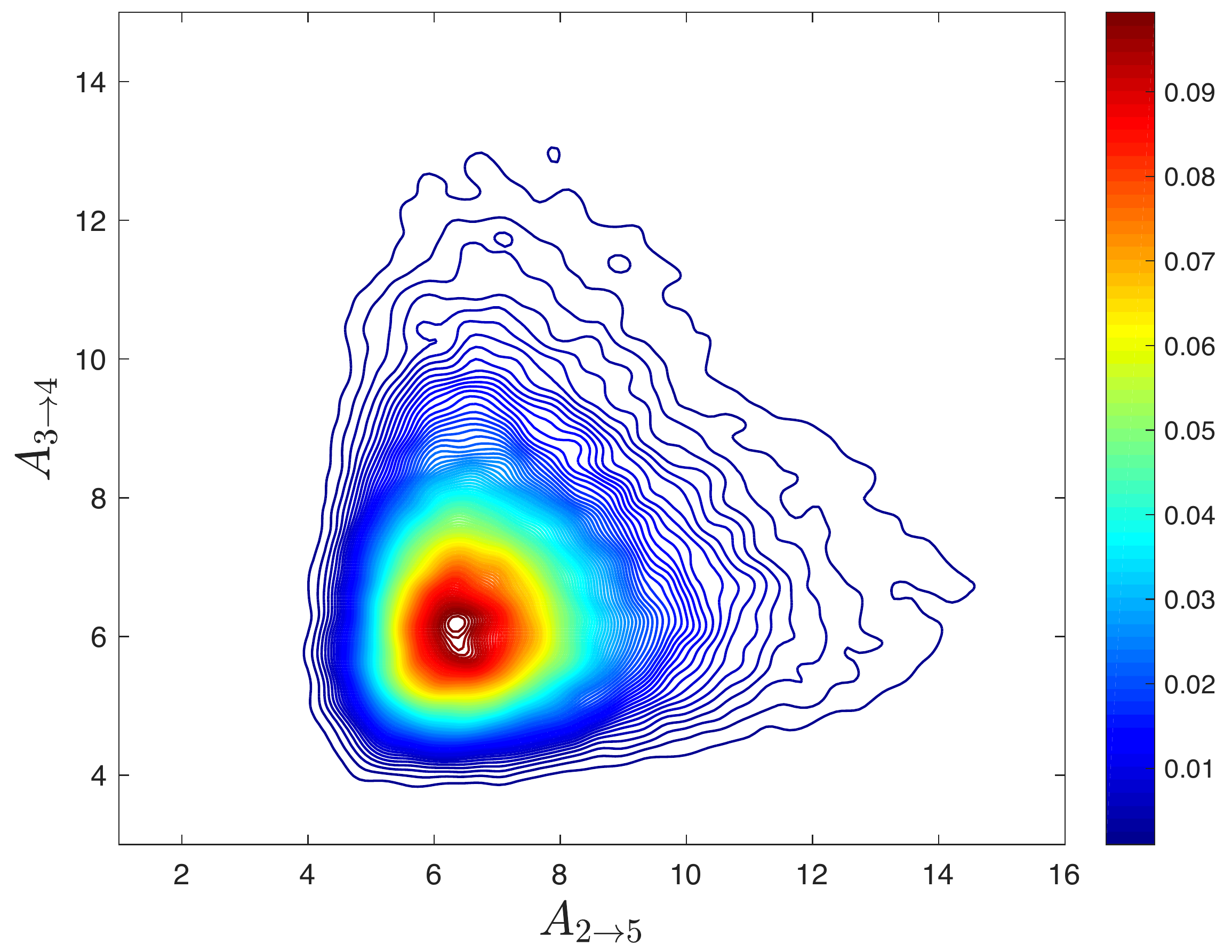}
  \caption{}
  \label{fig:7D-true-4}
  \end{subfigure}
\caption{A subset of 2-D contour projections of the true 7-D posterior probability distribution, $\pi_{\text{post}}(\Vector \theta)$, for the 6-node network problem, given a Gaussian prior and observed data from 20 experiments.}
\label{fig:7D-true}
\end{figure}

Fig.~(\ref{fig:7D-true}) shows 2-D contour projections of the true 7-D posterior probability distribution, $\pi_\text{post}(\Vector \theta)$, along the subset of dimensions labeled in the figure. Notice that, due to the higher noise level in the data, the range of support of the posterior probability this time is wider than the range of support (along the directions of the $A$ parameters) of the 6-D posterior probability in Section~\ref{sec:6D model}. Moreover, the location of the mode of the posterior distribution has been nudged to be a bit off from the underlying true value of the parameters. Comparing Figs.~(\ref{fig:7D-true-1}--\ref{fig:7D-true-3}) with Fig.~(\ref{fig:7D-true-4}), we can see, consistent with what we would expect, that the probability contours for parameters along different reaction pathways flare out more in the diagonal direction, whereas the probability contours for parameters that contribute to a shared reaction pathway are stretched in directions parallel to the axes. 
%%%%%%%%%%%%%%%%%%%%%%%%%%%%%%%% 
%%%%%%%%%%%%%%%%%%%%%%%%%%%%%%%%  

\begin{figure}[h!]
  \begin{subfigure}[t]{0.5\textwidth}
  \centering
  \includegraphics[scale=0.35]{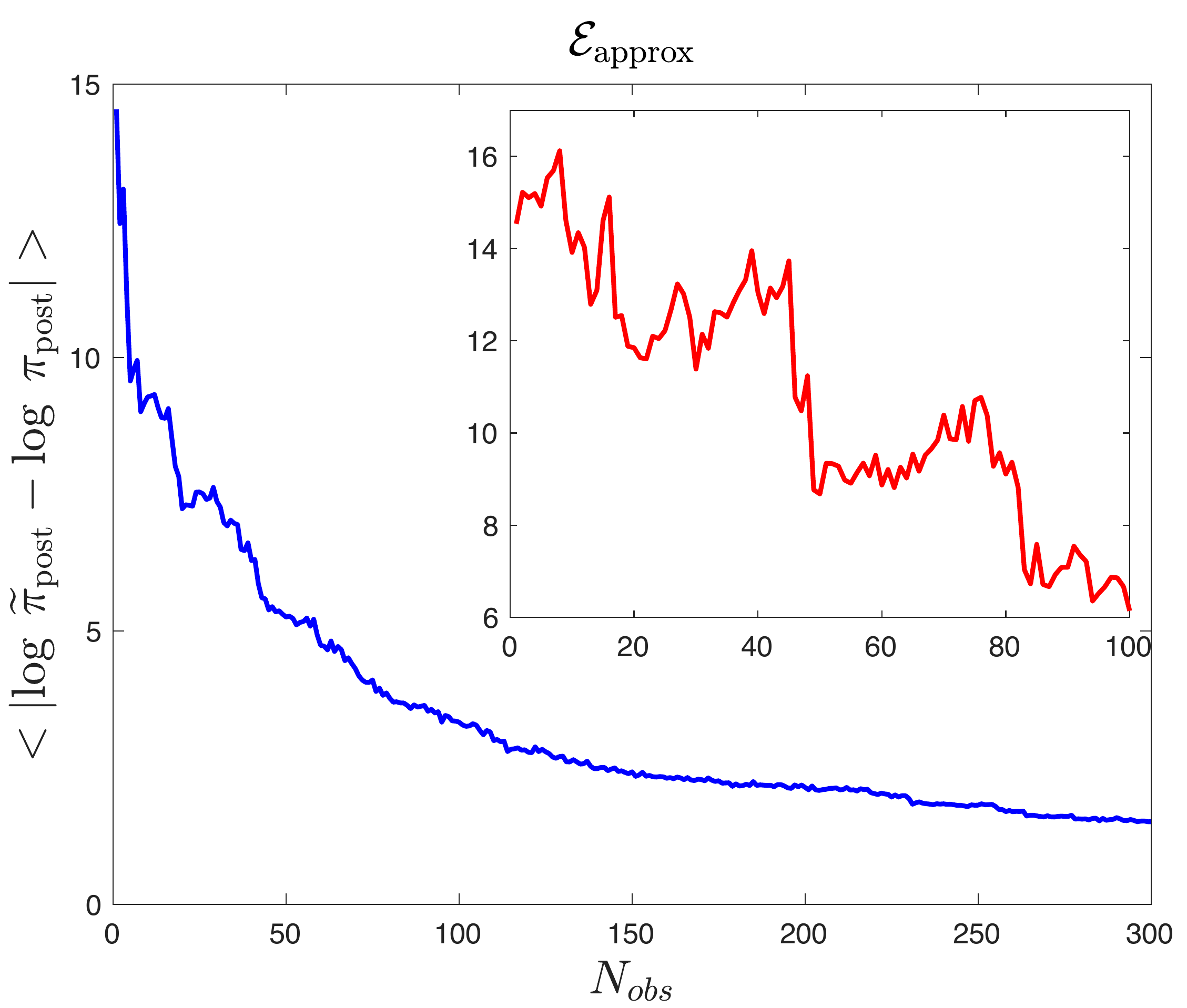}
  \caption{}
  \label{fig:7D-Egp}
  \end{subfigure}
  \begin{subfigure}[t]{0.5\textwidth}
  \centering
  \includegraphics[scale=0.35]{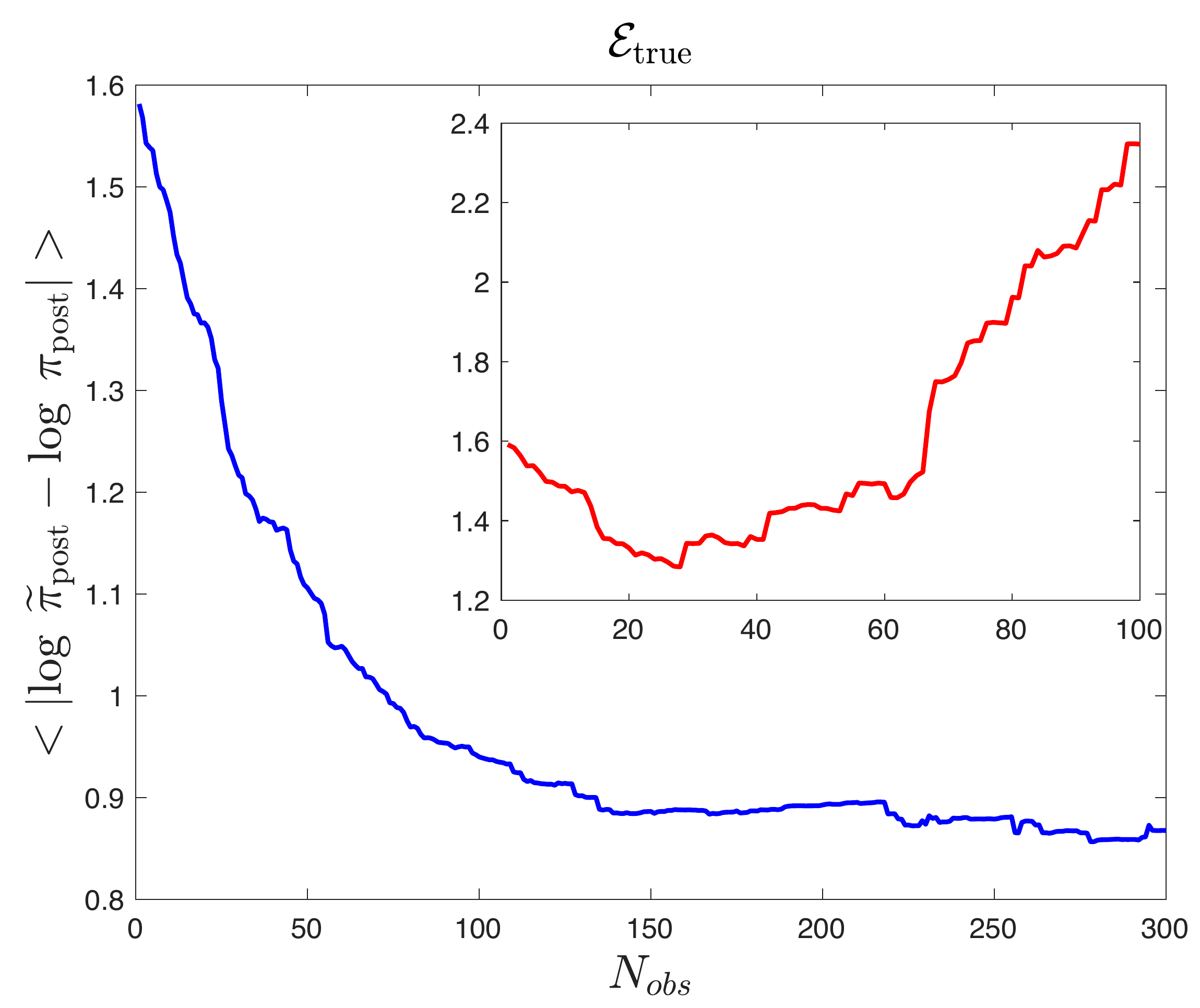}
  \caption{}
  \label{fig:7D-Etrue}
  \end{subfigure}
  \begin{subfigure}[t]{\textwidth}
  \centering
  \includegraphics[scale=0.35]{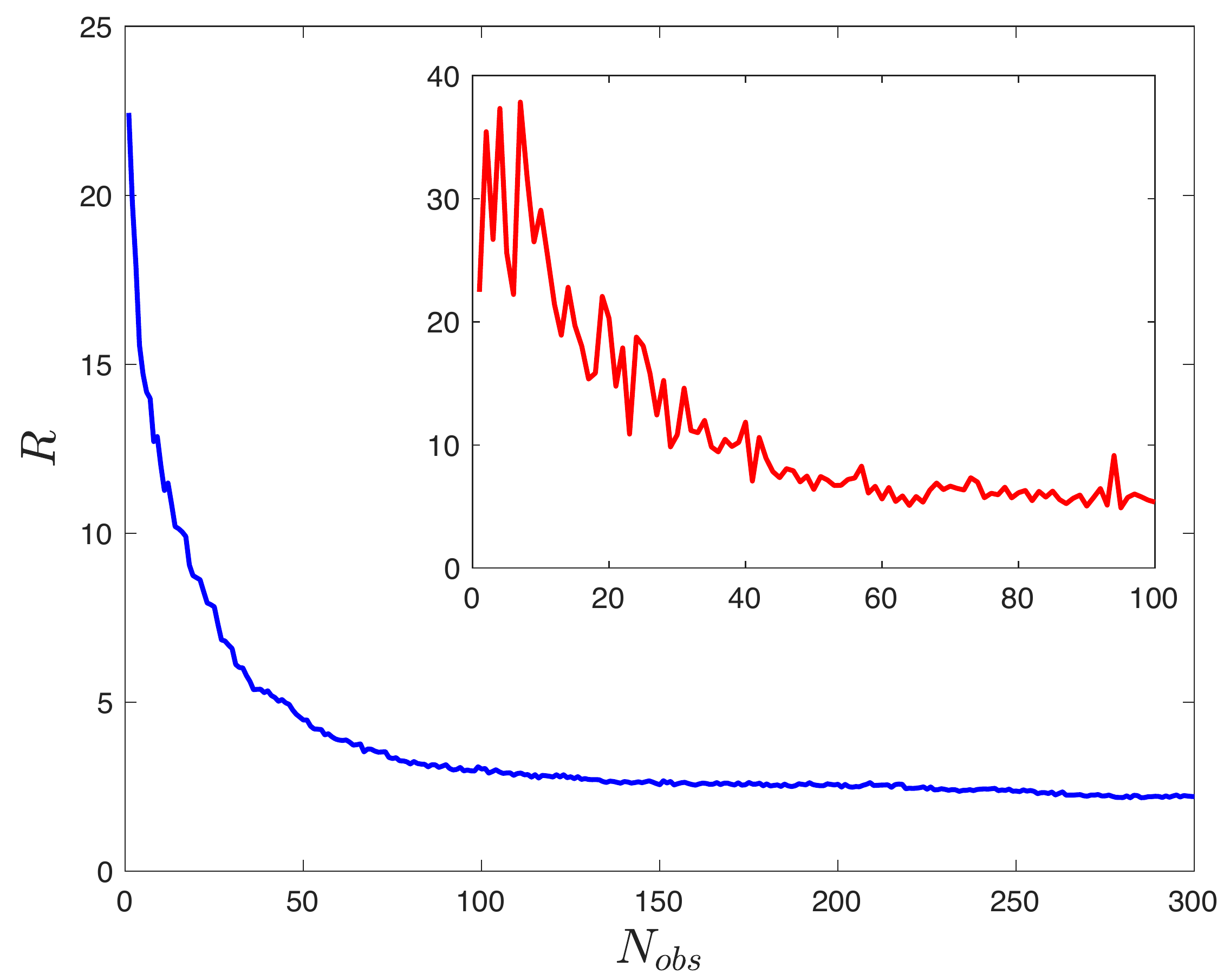}
  \caption{}
  \label{fig:7D-Rmeasure}
  \end{subfigure}
\caption{Evolution of the empirical average absolute errors and the $R$ measure of the 7-D surrogate surface as training points are added sequentially to the GP surface. Figure insets (red plots) correspond to  accuracy measures that result when training points are added using random selection, as opposed to the selection algorithm proposed in this paper.}
\label{fig:7D-error}
\end{figure}

Following the same methods used in Section~\ref{sec:6D model} to construct the surrogate surface approximation, $\widetilde{\pi}_\text{post}(\Vector \theta)$, and skipping over the qualitative comparison of the probability contours for the sake of brevity, we move on to check the quality of our surrogate approximation using the accuracy measures described earlier. Figs.~(\ref{fig:7D-Egp})--(\ref{fig:7D-Rmeasure}) show the evolution of $\mathcal E_\text{approx}$, $\mathcal E_\text{true}$, and the $R$ measure, respectively, as we sequentially add training points to the GP-augmented surface. Again, overall monotonic decreasing trends are observed as more training points are sequentially fed to the GP surface, with the rate of decrease being highest at the beginning and becoming more gradual later on. However this time, as is evident from initial values of the absolute errors and R measure at $N_{obs} = 0$, the initial GP-augmented surface does not start out as being a close approximation to the true posterior surface. In fact, almost an order of magnitude reduction in the absolute error and the R measure is achieved by the end of the iterations ($R(0) \approx 22, R(300) \approx 2.2 \, ; \mathcal E_\text{approx}(0) \approx 14.5, \mathcal E_\text{approx}(300) \approx 1.5 \, ; \mathcal E_\text{true}(0) \approx 1.6, \mathcal E_\text{true}(300) \approx 0.8$). 

%%%%%%%%%%%%%%%%%%%%%%%%%%%%%%%%
%%%%%%%%%%%%%%%%%%%%%%%%%%%%%%%%

To rule out the possibility that our training algorithm is not any better than a simple random selection of training points, we repeated the 7-D exercise above using the same methods and constraints, but instead of relying on Eq.~(\ref{eq:var_approx}) as a criterion for selecting training points at each iteration, we sampled $\widetilde{\pi}_\text{post}(\Vector \theta)$ and randomly selected one of the (after burn-in) sample points to be our next training data point. The resulting accuracy measures are shown in the insets of Figs.~(\ref{fig:7D-Egp})--(\ref{fig:7D-Rmeasure}). Contrary to the trends observed when using the training algorithm, the average absolute error measures using this random selection technique failed to converge, as can be seen from the inset in Fig.~(\ref{fig:7D-Etrue}). The eventual increasing trend observed in $\mathcal E_\text{true}$, despite the overall decreasing trends in $\mathcal E_\text{approx}$ and $R$, is an indication that the GP has crashed. This is because the GP is tending towards correctly capturing the low probability regions, which gives rise to the decrease in $\mathcal E_\text{approx}$, at the expense of misrepresenting the high probability regions, which causes the increase in $\mathcal E_\text{true}$. The combination of these two effects leads to a decrease in $R$, even though the surrogate surface is progressively deviating from the actual true surface. This justifies our earlier cautioning against relying on any one of the accuracy measures exclusively. The collapse of the GP when training points are selected randomly, underscores the importance of correctly weighting the potential observation points using a utility measure similar to the one proposed in Section~\ref{sec:utility measure}.

\bigskip\bigskip
\section{Conclusion}
\label{sec:conc}

In this paper, we sought to construct a surrogate approximation of the posterior probability distribution that results during the Bayesian inference of model parameters for expensive forward models, in order to tackle the bottleneck of having to solve the forward model multiple times during MCMC sampling. The surrogate surface was built iteratively using a stationary, isotropic GP model with fixed (unoptimized) kernel parameters. To help maximize the efficiency of the GP training process, an algorithm was developed which seeks, at each iteration, the optimal data point to add to the GP training set using a point-wise probability-weighted utility measure. 

Motivated by the inference of reaction rate parameters in combustion applications, the capability of the algorithm to recover 2-D, 6-D, and 7-D posterior probability distributions was tested on 3-node and 6-node network models. Starting with an initial Gaussian approximation of the posterior, and despite employing a possibly non-optimal GP model, the algorithm was able to successfully re-construct the true 2-D posterior distributions and achieve almost an order of magnitude increase in accuracy after about a 100 iterations in the higher dimensional cases. When comparing our sequential learning algorithm to that of a passive (random) learner in the 7-D scenario, the latter crashed after about 30 iterations, which is testament to the importance of judiciously selecting the GP training points when seeking to create surrogates to posterior probability distributions. The present experiences warrant further testing and development of the current sequential learning algorithm, particularly with regard to incorporating more sophisticated GP models and selecting multiple new training points at once.

\bigskip\bigskip
\section*{Acknowledgments}

This work was supported by the U.S.\ Department of Energy, Office of Science, Office of Advanced Scientific Computing Research, Applied Mathematics Program under contract DE-AC02-05CH11231. This research used resources of the National Energy Research Scientific Computing Center, a DOE Office of Science User Facility supported by the Office of Science of the U.S.\ Department of Energy under contract DE-AC02-05CH11231.

\bigskip\bigskip

\bibliographystyle{unsrt}
\bibliography{references}

\end{document}